\documentclass[10pt,twocolumn,twoside]{IEEEtran}

\usepackage{times}
\usepackage{epsfig}
\usepackage{graphicx}
\usepackage{amsmath}
\usepackage{amssymb}
\usepackage{url}
\usepackage{multirow}
\usepackage[table,xcdraw]{xcolor}
\usepackage{threeparttable}

\usepackage{bbding}
\usepackage{graphicx}
\usepackage{subfigure}
\usepackage{bbm}
\usepackage{float}
\usepackage{algorithm}
\usepackage{algpseudocode}

\usepackage{booktabs}
\makeatletter
\renewcommand{\maketag@@@}[1]{\hbox{\m@th\normalsize\normalfont#1}}%
\makeatother
\usepackage[table,xcdraw]{xcolor}
\definecolor{mygray}{gray}{.9}
\usepackage{multirow}

\usepackage{expl3}
\ExplSyntaxOn
\newcommand\latinabbrev[1]{
  \peek_meaning:NTF . {% Same as \@ifnextchar
    #1\@}%
  { \peek_catcode:NTF a {% Check whether next char has same catcode as \'a, i.e., is a letter
      #1.\@ }%
    {#1.\@}}}
\ExplSyntaxOff

\ifCLASSINFOpdf
\else
\fi

\begin{document}

% \markboth{Journal of \LaTeX\ Class %Files,~Vol.~6, No.~1, January~2007}
\title{SMTrack: State-Aware Mamba for Efficient Temporal Modeling in Visual Tracking}

\author{Yinchao~Ma,
        Dengqing~Yang,
        Zhangyu~He,
        Wenfei~Yang,
        Tianzhu~Zhang

% \thanks{Manuscript received *** **, 2016; revised *** **, 2016; accepted ***
% **, 2017. This work was supported in part by the National Natural Science
% Foundation of China under Grant 61225009, Grant 61432019, Grant 61572498, 
% Grant 61532009,  and Grant 61572296, and in part by the Importation and Development of
% High-Caliber Talents Project of Beijing Municipal Institutions under Grant
% IDHT20140224.}

% \IEEEcompsocitemizethanks{Junyu Gao, Tianzhu Zhang, Xiaoshan Yang and Changsheng Xu are with National Lab of Pattern Recognition, Institute of
% Automation, Chinese Academy of Sciences, Beijing 100190, P. R. China (e-mail: gaojunyu2015@ia.ac.cn;
% tzzhang@nlpr.ia.ac.cn;
% xiaoshan.yang@nlpr.ia.ac.cn;
% csxu@nlpr.ia.ac.cn).}
}




\markboth{IEEE Transactions on Image Processing,~Vol. XX, ~No. XX,~Aug 202X}
{Ma \MakeLowercase{\textit{et al.}}: Learning Discriminative Features for Visual Tracking via Scenario Decoupling}



\maketitle

\begin{abstract}
Visual tracking aims to automatically estimate the state of a target object in a video sequence, which is challenging especially in dynamic scenarios.
Thus, numerous methods are proposed to introduce temporal cues to enhance tracking robustness.
However, conventional CNN and Transformer architectures exhibit inherent limitations in modeling long-range temporal dependencies in visual tracking, often necessitating either complex customized modules or substantial computational costs to integrate temporal cues.
Inspired by the success of the state space model, we propose a novel temporal modeling paradigm for visual tracking, termed State-aware Mamba Tracker (SMTrack), providing a neat pipeline for training and tracking without needing customized modules or substantial computational costs to build long-range temporal dependencies.
It enjoys several merits.
First, we propose a novel selective state-aware space model with state-wise parameters to capture more diverse temporal cues for robust tracking.
Second, SMTrack facilitates long-range temporal interactions with linear computational complexity during training.
Third, SMTrack enables each frame to interact with previously tracked frames via hidden state propagation and updating, which releases computational costs of handling temporal cues during tracking.
Extensive experimental results demonstrate that SMTrack achieves promising performance with low computational costs.

%
% Visual tracking aims to automatically estimate the state of a target object in a video sequence, which is challenging especially in dynamic scenarios. Thus, numerous methods are proposed to introduce temporal cues to enhance tracking robustness. However, conventional CNN and Transformer architectures exhibit inherent limitations in modeling long-range temporal dependencies in visual tracking, often necessitating either complex customized modules or substantial computational costs to integrate temporal cues. Inspired by the success of the state space model, we propose a novel temporal modeling paradigm for visual tracking, termed State-aware Mamba Tracker (SMTrack), providing a neat pipeline for training and tracking without needing customized modules or substantial computational costs to build long-range temporal dependencies. It enjoys several merits. First, we propose a novel selective state-aware space model with state-wise parameters to capture more diverse temporal cues for robust tracking. Second, SMTrack facilitates long-range temporal interactions with linear computational complexity during training. Third, SMTrack enables each frame to interact with previously tracked frames via hidden state propagation and updating, which releases computational costs of handling temporal cues during tracking. Extensive experimental results demonstrate that SMTrack achieves promising performance with low computational costs.

\end{abstract}


% \fi
\IEEEpeerreviewmaketitle

\section{Introduction}
\label{sec:intro}

Visual tracking is a fundamental task in the field of computer vision, aimed at automatically localizing a given reference object within a video sequence.~\cite{yilmaz2006object}.
It has been successfully applied in a variety of real-world applications, including human-computer interaction, visual surveillance, robotics, and autonomous driving~\cite{zhang2013robust,liu2015fashion}.
Although significant progress has been made in recent years, visual tracking still encounters several challenges, including occlusion, deformation, dynamic backgrounds and so on~\cite{OTB2015}.

\begin{figure*}[t]
    \centering
    \includegraphics[width=1.0\linewidth]{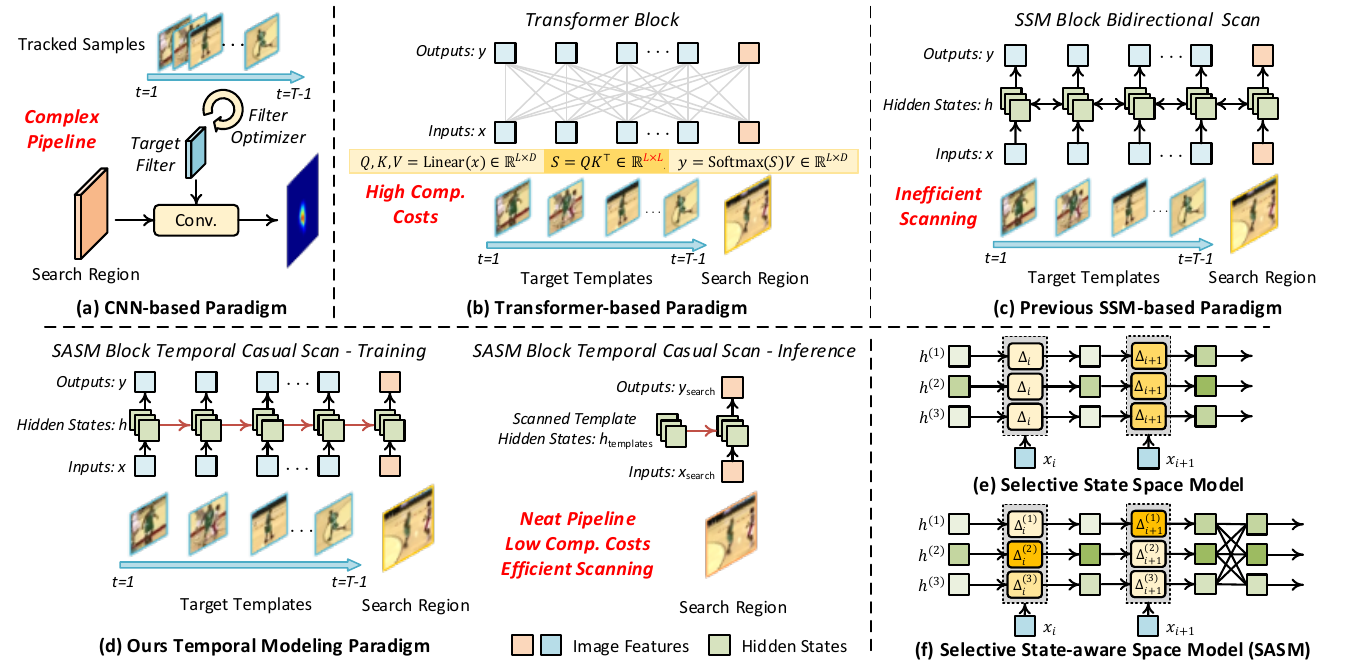}
    \caption{Classical temporal modeling paradigm with different network architecture for visual tracking. Here, \textit{Comp.} is short for \textit{Computational}.
    (a) CNN-based paradigm uses well-designed optimizers to train a target filter with online tracked samples, leading to complex pipelines.
    (b) Transformer-based paradigm concatenates temporal cues ($e.g.$ dynamic templates) into inputs to build dependencies, which needs quadratic computational costs.
    (c) The previous SSM-based paradigm adopts bidirectional SSM to introduce more templates, whose cross-frame bidirectional scan requires templates to be scanned repeatedly for each frame during tracking, limiting efficiency.
    (d) Our temporal modeling paradigm introduces temporal cues via temporal causal scanning with linear complexity for training and builds long-range interactions via state propagation and updating for inference without needing scanning templates repeatedly, which greatly releases the computational costs of temporal modeling and provides a neat tracking pipeline.
    (e) The previous selective state space model employs a state-shared timescale as a gating mechanism, making features uniformly ignored or considered for all state updating.
    (f) Our selective state-aware space model assigns state-wise timescales for hidden states, enabling them to capture more diverse temporal cues.
    }
    \label{fig:intro}
\end{figure*}

To overcome the above challenges of scenario variations, temporal modeling is attracting more attention to enhance tracking robustness.
Numerous trackers with temporal modeling have been designed based on advanced network architectures, including CNNs~\cite{DiMP,ATOM,ECO} and Transformers~\cite{yan2021learning,cui2022mixformer,VideoTrack}.
\textbf{1) Temporal modeling based on CNN}. 
% 场景信息对于跟踪模型判别目标的位置至关重要
As shown in Figure~\ref{fig:intro}(a), most CNN-based trackers introduce temporal cues by training a target filter (correlation filter~\cite{CCOT,KCF,ECO} or convolution filter~\cite{ATOM,DiMP,DROL,Ocean}) using tracked samples to seek a highly responsive position in the search region, which desire a well-designed filter optimizer to speed up the online training process.
Besides, some CNN-based trackers~\cite{SiamRCNN,mayer2021learning,KYS} seek to build temporal interactions by hand-crafted cross-frame association scores~\cite{SiamRCNN,Dsiamese,DaSimese} or association networks~\cite{SiamRCNN,KYS,UpdateNet}, which lead to complex network structures and numerous computational costs.
\textbf{2) Temporal modeling based on Transformer}. 
Many Transformer-based trackers~\cite{yan2021learning,High-TransT,cui2022mixformer,APMT,TMT} introduce temporal cues by concatenating target templates of tracked frames into inputs, as shown in Figure~\ref{fig:intro}(b).
% each feature ne
However, the attention mechanisms in Transformer build interactions by similarities between all features, which leads to quadratic complexity of sequence length, making the computational costs increase rapidly as the number of target templates increases.
% 为了避免误差积累以及模板选择机制
Beyond the sparse temporal cues of target templates, recent works~\cite{AQATrack,ODTrack,EVPTrack} seek to build associations between consecutive frames.
They introduce customized modules to propagate temporal cues, such as temporal tokens~\cite{ODTrack,ROMTrack}, visual prompts~\cite{EVPTrack,HIPTrack} and so on\cite{VideoTrack,ARTrackV2}.
However, customized modules inevitably bring complex network structures and higher computational costs.
Also, these trackers commonly desire to sample multiple search frames and predict them sequentially for temporal cue propagation during training, which complicates the training pipeline.

Based on the discussions above, we can find that CNNs and Transformers exhibit inherent limitations in modeling long-range temporal dependencies in visual tracking, necessitating either complex customized modules or substantial computational costs during tracking.
Recently, the success of the state space model (SSM), represented by Mamba~\cite{Mamba}, inspires us to explore a new tracking paradigm.
Mamba strikes a balance between facilitating global interactions and maintaining linear complexity.
Meanwhile, Mamba enables each element to interact with any of the previously scanned samples through a compressed hidden state, which greatly improves the inference efficiency of long sequences in natural language processing~\cite{Mamba}.
In computer vision, many methods~\cite{Vim,VideoMamba,VMamba,SegMamba,MTMamba} introduce multi-directional SSMs to build global interactions of 2D images with linear complexity, which achieve competitive performance against attention-based methods, demonstrating the superiority of SSM in vision tasks.
A natural question arises: \textbf{Can we design a neat temporal modeling paradigm based on SSMs for visual tracking?}

SSMs have the potential to efficiently model long-range temporal dependencies.
However, there are two main issues that need to be considered for visual tracking.
\textbf{(1) Diverse Temporal Cues for Robust Tracking.}
Target templates contain diverse cues (such as targets, backgrounds, and distractors), which are essential for enabling trackers to not only localize the target object but also suppress backgrounds and distractors for robust tracking~\cite{KYS,mayer2021learning}.
However, as shown in Figure~\ref{fig:intro}(e), the popular SSM, Mamba~\cite{Mamba}, employs a state-shared timescale parameter $\Delta$ as a gating mechanism, making features uniformly ignored or considered for all states updating.
Such a design limits the ability of hidden states to capture diverse cues.
%
% It is better for hidden states to capture diverse temporal cues for visual tracking.
%
% Besides, hidden states are learned independently, making them unable to be optimized from a global perspective to capture more comprehensive cues.
%
\textbf{(2) Temporal Causality for Visual Tracking.}
Recently, MambaVT~\cite{MambaVT} designs an SSM-based bidirectional scan framework to build dependencies of temporal cues for RGB-T tracking, as shown in Figure~\ref{fig:intro}(c).
However, its cross-frame bidirectional scan requires target templates to be repeatedly scanned with the future information ($e.g.$ search region) in each frame, greatly impairing tracking efficiency.
We argue that it is better to design a new scanning mechanism that decouples templates from future information, allowing the search region to directly interact with templates via hidden states without repetitive scanning of templates.

Motivated by the above discussions, we propose a novel temporal modeling paradigm for visual tracking, termed \textbf{S}tate-aware \textbf{M}amba Tracker (SMTrack), providing a neat pipeline for training and tracking with temporal cues.
Firstly, we propose a selective state-aware space model (SASM), as shown in Figure~\ref{fig:intro}(f).
SASM employs state-wise timescale parameter $\Delta$, enabling hidden states to capture more diverse temporal cues.
%%%%%%%%%%%%%%%%%%%%%%%%%%%%%%%%%%%%%%%%%%%%%%%
Meanwhile, SASM introduces interactions of hidden states to build dense dependencies between them.
% so that they can be optimized from a global perspective.
%
Secondly, we build SMTrack based on SASM blocks with temporal causal scanning, as shown in Figure~\ref{fig:intro}(d).
On the one hand, SMTrack can introduce more temporal cues into model training with linear complexity.
On the other hand, temporal causal scanning enables the search region to interact with scanned templates via hidden states without the need for repeated scanning of templates during tracking.
Meanwhile, we keep bidirectional scanning intra-frame for spatial-aware feature learning.
Further, SMTrack can easily integrate new temporal cues via scanning templates of tracked frames to update the hidden states without needing extra module designing. 

In summary, the main contributions of this work are:
(1) We propose a new temporal modeling paradigm for visual tracking, SMTrack, providing a neat pipeline for training and tracking with temporal cues.
SMTrack facilitates interaction between search region and temporal cues via hidden state propagation and updating without needing customized modules or substantial computational costs.
(2) We propose a novel selective state-aware space model (SASM) that employs state-wise timescale parameters to capture more diverse temporal cues for robust tracking.
%
% Moreover, SASM introduces interactions among hidden states, enabling state optimization from a global perspective.
%
(3) Extensive experimental results on tracking benchmarks demonstrate that SMTrack achieves superior performance with low computational costs.
% We conduct extensive experiments on tracking benchmarks showing the efficacy of SMTrack.

%-------------------------------------------------------------------------

\section{Related Works}

\subsection{CNN-based Trackers}
CNN-based trackers can generally be divided into two categories, including Discriminative Correlation Filter based (DCF-based) trackers~\cite{CCOT,KCF,ECO,ATOM,DiMP,DROL,Ocean} and Siamese-based trackers~\cite{SiameseFC,TIP5,TIP2,SiameseRPN,SiamRPNplusplus,SiamFC++,DaSimese}.
The basic idea of DCF-based methods is to train a correlation filter~\cite{CCOT,KCF,TIP4,TIP3} or a convolution filter~\cite{ATOM,DiMP,Ocean,DROL} using online tracked samples to seek a highly responsive position in the search region.
They desire a well-designed optimizer to speed up the online training process, such as conjugate gradient optimizer~\cite{ECO,ATOM,CCOT} and steepest descent optimizer~\cite{DiMP,DROL,Ocean}.
Siamese-based trackers~\cite{SiameseFC,SiameseRPN,SiamRPNplusplus} extract features for both target template and search region with the Siamese networks, and estimate object state by matching their features, which is essentially a pair-wise matching framework.
To overcome the challenge of dynamic scenarios, some Siamese-based trackers~\cite{SiamRCNN,DaSimese,SiamRCNN,KYS} further introduce temporal cues by hand-crafted cross-frame association scores~\cite{SiamRCNN,Dsiamese,DaSimese} or association networks~\cite{SiamRCNN,KYS,UpdateNet}, which leads to complex network structure.
Unlike CNN-based trackers, SMTrack can naturally build interactions between the search region and temporal cues via state propagation and updating without needing customized modules.

\subsection{Transformer-based Trackers}
Thanks to its global modeling capacity and parallel processing capabilities,
Transformer~\cite{2017Attention} has been well-developed in visual tracking.
Pioneering works like TransT~\cite{TransT} and OSTrack~\cite{OSTrack} adopt Transformers to build global and non-linear interactions between target template and search region, achieving superior performance.
To adapt to target variations, many trackers~\cite{yan2021learning,High-TransT,cui2022mixformer,APMT,TIP1} incorporate temporal cues by directly concatenating dynamic templates into the input.
However, the quadratic complexity of attention mechanisms makes it challenging to retain more dynamic templates.
To construct more robust sequence-level temporal associations, recent works~\cite{HIPTrack,ROMTrack,VideoMamba} design various modules to propagate temporal cues in consecutive frames, such as temporal tokens~\cite{ODTrack,ROMTrack}, visual prompts~\cite{EVPTrack,HIPTrack}, autoregressive queries~\cite{AQATrack} and generated template~\cite{ARTrackV2}, then integrate them with search region features via fusion modules.
However, customized modules inevitably bring higher computational costs and complex network structures.
Moreover, these trackers commonly desire to sample multiple search frames and predict them sequentially for temporal cue propagation during training, which complicates the training pipeline.
Differently, SMTrack can directly introduce more dynamic templates to build long-range temporal interactions with linear complexity for training and can integrate temporal cues via hidden states propagation and updating for tracking, which greatly releases the computational costs of temporal modeling.

\subsection{State Space Models}
State Space Models~\cite{S4,S5,H3,SSM1,SSM2,reviewer1,reviewer2,reviewer3}, drawing inspiration from continuous systems, have become promising approaches for sequence modeling.
Recently, the selective state space models, represented by Mamba~\cite{Mamba}, employ an input-driven selection mechanism to model long-range dependencies linearly with sequence length, showing competitive performance against Transformer in natural language processing.
In computer vision, Vim~\cite{Vim} and VMamba~\cite{VMamba} first introduce multi-directional SSMs to build global interactions of 2D images, which achieve promising performance.
Thanks to the linear computational complexity of SSM, numerous methods adopt SSM to solve tasks that need long sequence interactions, such as video understanding~\cite{VideoMamba,MTMamba,ViVim}, medical image segmentation~\cite{SegMamba,SwinUMamba,VM-UNet}, voxel modeling~\cite{VoxelMamba} and so on~\cite{RSMamba,ChangemMamba}.
In tracking tasks, MambaTrack employs Mamba as a motion model to predict complex motion patterns for MOT.
MambaVT~\cite{MambaVT} designs an SSM-based bidirectional scan framework to build dependencies with temporal cues for RGB-T tracking.
Although MambaVT introduces temporal cues with linear complexity, its cross-frame bidirectional scanning requires target templates to be updated by hidden states of the search region, leading to repetitive scanning of templates for each frame during tracking.
Different from MambaVT, SMTrack seeks to keep the temporal causality across frames so that SMTrack can build interactions between each frame and scanned templates through a compressed hidden state without repeated scanning during tracking.
Also, beyond Mamba, we further propose a selective state-aware space model with state-wise timescale parameters to capture more diverse temporal cues for robust tracking.

%-------------------------------------------------------------------------
%
%
%
\begin{figure*}
    \centering
    \includegraphics[width=1.0\linewidth]{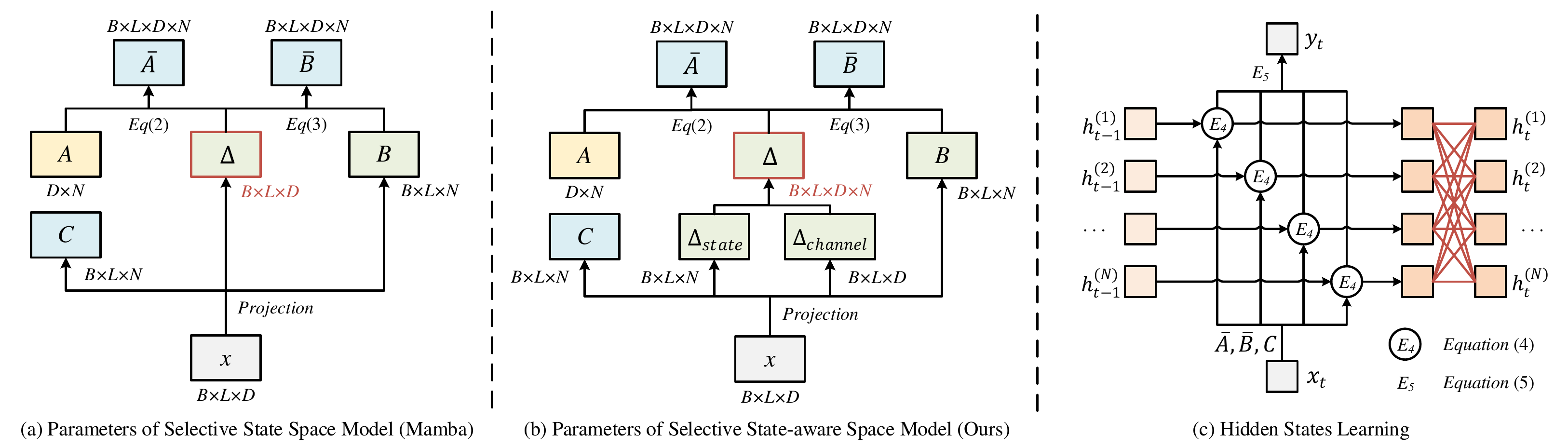}
    \caption{
    Parameters of different state space models.
    $B,L,D,N$ are $batch$, $sequence$, $channel$, $state$ dimensions respectively.
    (a) shows the parameter generation of the selective state space model.
    The timescale parameter $\Delta$ is state-shared, which limits the diverse interactions between different hidden states and image features.
    (b) shows the parameter generation of our selective state-aware space model (SASM).
    We utilize a state-wise timescale parameter $\Delta$, enabling different hidden states to capture diverse temporal cues, such as targets, backgrounds and distractors. 
    (c) shows the hidden states learning of SASM, which introduces interactions among hidden states after each image (shown in red lines) to build dense dependencies between them.
    }
    \label{fig:SASM}
\end{figure*}

\section{Method}
In this section, we first introduce the preliminaries of the advanced selective state space model.
The following subsection introduces the details of our proposed selective state-aware space model (SASM).
In the last subsection, we present the novel temporal modeling paradigm of SMTrack.

\subsection{Preliminaries}

\textbf{State Space Models.}
State Space Models are formulated from linear time-invariant continuous systems that map 1D sequence $x(t) \in \mathbb{R}^{L} \rightarrow y(t) \in \mathbb{R}^{L}$ via a hidden state $h(t) \in \mathbb{R}^{N}$.
They typically employ linear ordinary differential equations (ODEs) to build the mapping. Formally,
\begin{equation}\label{eq:ssm}
\begin{split}
  &h'(t) = {A}h(t) + {B}x(t)  \\
  &y(t)  = {C}h(t),
\end{split}
\end{equation}
where $A \in \mathbb{R}^{N \times N}$ denotes evolution parameter, $B\in \mathbb{R}^{N \times 1}, C \in \mathbb{R}^{1 \times N}$ are projection parameters.

\textbf{Discretization.}
To integrate into deep learning algorithms, the continuous ODE is approximated through discretization in modern SSMs~\cite{Mamba,S4}, which is typically accomplished using the zero-order hold (ZOH) method,
\begin{gather}
\bar{A} = e^{\Delta A},\label{equ:2} \\ 
\bar{B} = ( e^{\Delta A} - I)A^{-1}B \approx \Delta B, \label{equ:3} \\ 
h_t = \bar{A}h_{t-1} + \bar{B}x_t, \label{equ:4} \\ 
y_t = {C}h_t,
\end{gather}
where $\bar{A}$ and $\bar{B}$ are the discrete counterparts of parameters $A$ and $B$, $\Delta$ is a timescale parameter, $h_t,h_{t-1}$ denote the discrete hidden states at various time steps.

\textbf{Selective State Space Model.}
Recently, Mamba~\cite{Mamba} further implements a selective state space model with a selective scan mechanism, in which the parameters $B \in \mathbb{R}^{B \times L  \times N}$, $C \in \mathbb{R}^{B \times L  \times N}$, $\Delta \in \mathbb{R}^{B \times L  \times D}$ arise from the input data $x \in \mathbb{R}^{B \times L  \times D}$, enabling the ability of content-aware reasoning and adaptive weight modulation.

\subsection{Selective State-aware Space Model}
The popular selective state space model in Mamba~\cite{Mamba} strikes a balance between facilitating global interactions and maintaining linear complexity.
%
% There are also many methods~\cite{Vim,VideoMamba,VMamba,SegMamba,MTMamba} in computer vision introducing selective state space model to build global interactions of images, achieving superior performance.
%
However, as shown in Figure~\ref{fig:SASM}(a), the selective state space model employs a state-shared timescale parameter $\Delta$ as a gating mechanism to control which features participate in state updating as Equation~(\ref{equ:2}-\ref{equ:4}) does.
Such a state-shared design makes features uniformly ignored or considered for all states, thereby limiting the ability of hidden states to capture diverse cues for visual tracking.
% may limit the diverse interactions between different hidden states and image features.
% 

\begin{figure}[t]
    \centering
    \includegraphics[width=1.0\linewidth]{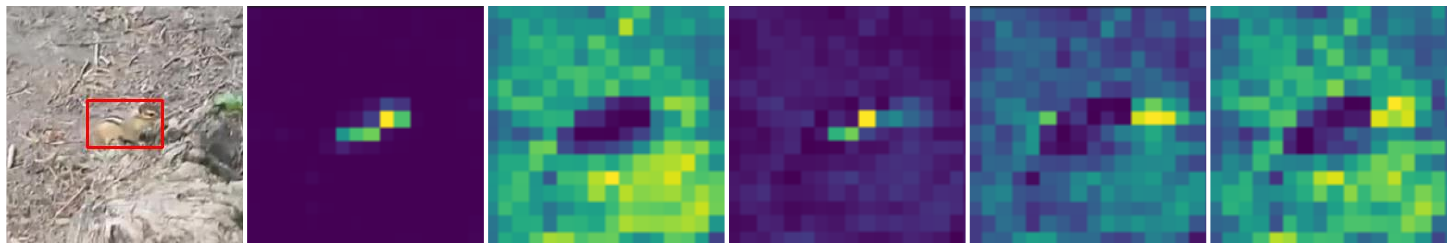}
    \caption{The first column is the target template. We sum the timescale parameter $\Delta$ in SASM along the channel dimension and reshape it into the 2D map to show its response at different template regions for each hidden state. SASM enables different hidden states to freely capture diverse scenario cues, such as targets, backgrounds or distractors.
    }
    \label{fig:delta}
\end{figure}

To overcome the above limitations, we design a selective state-aware space model (SASM) with a state-wise timescale parameter $\Delta$, as shown in Figure~\ref{fig:SASM}(b).
We extract state-wise and channel-wise timescale parameters $\Delta_{state}$ and $\Delta_{channel}$ separately via linear projections and then add them together via broadcasting. Formally,
\begin{gather}
    \Delta_{state} = {\rm Linear_{s}}(x), \Delta_{channel} = {\rm Linear_{c}}(x),\\
    \Delta = \sigma({\rm Broadcast}(\Delta_{state}) + {\rm Broadcast}({\Delta_{channel}})),
\end{gather}
where $\sigma(\cdot)$ denotes softplus operation, $\Delta_{channel}$ controls whether each channel of features should be ignored or considered as Mamba~\cite{Mamba} does.
$\Delta_{state}$ controls whether features should be ignored or considered for each state.
Here, we broadcast $\Delta_{state}$ in $3^{rd}$ dimension and $\Delta_{channel}$ in $4^{th}$ dimension, making them have the same shape of $(B, L, D, N)$.
% 添加delta可视化与分类结果图
By this design, different hidden states can freely capture diverse scenario cues on the target template, such as targets, backgrounds or distractors, as shown in Figure~\ref{fig:delta}.
Thereby, the scanned hidden states can retain more diverse temporal cues.
More state-wise parameter designs and visualizations are discussed in Section~\ref{sec:exp}.

Besides, the selective state space model learns hidden states independently of each other, as shown in Figure~\ref{fig:SASM}(c) black lines.
We further introduces interactions of hidden states to build dense dependencies between them, as shown in Figure~\ref{fig:SASM}(c) red lines.
Experiments show that this design helps SASM to more learn effective features for robust tracking.
%
% To overcome the above limitation, we introduce interaction among hidden states after scanning each image, as shown in Figure~\ref{fig:SASM}(c) red lines, achieving the exchange of information so that they can be optimized from a global perspective.
%
Formally,
\begin{gather}
    h = {\rm Linear_{up}}({\rm Linear_{down}}(h^{\prime})),
\end{gather}
where $h^{\prime}\in \mathbb{R}^{B\times D\times N}$ denotes the last hidden states of each image.
% $B,D,N$ are $batch$, $channel$, $state$ dimensions respectively.
%
$Linear_{down}(\cdot)$ decrease the dimensions of state to $N/4$.
$Linear_{up}(\cdot)$ increase the dimensions of state to $N$.
For simplicity, SASM can be generally expressed as,
\begin{gather}
    (F_{out}, h_{last})={\rm SASM}(F, h_{init}).
\end{gather}
Here, $F,F_{out}\in \mathbb{R}^{B\times L\times D}$ are the input and output features, $h_{init},h_{last}\in \mathbb{R}^{B\times D\times N}$ are initial and last hidden states.

\begin{figure*}[t]
    \centering
    \vspace{-3mm}
    \includegraphics[width=0.95\linewidth]{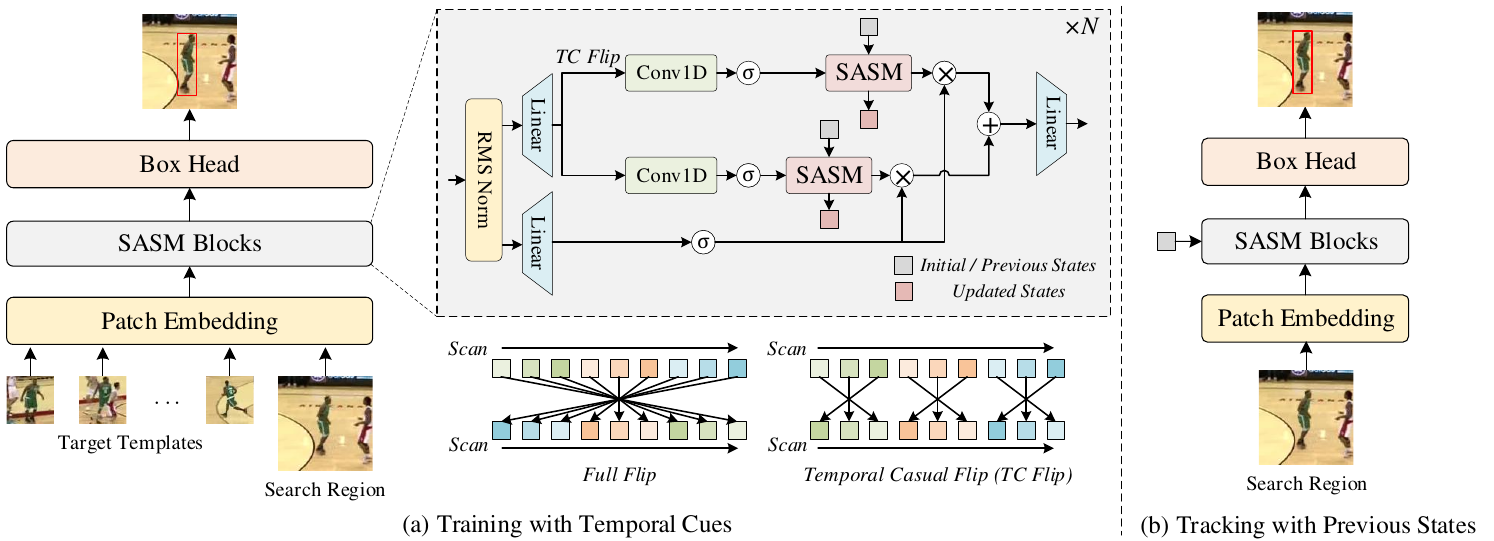}
    \caption{The architecture of SMTrack. 
    (a) SMTrack provides a neat pipeline to introduce temporal cues with linear complexity for training. 
    Meanwhile, we design a temporal causal flip method for temporal causal scanning, allowing the scanning process to maintain causality between frames.
    (b) SMTrack enables the search region to interact with any of the previously scanned frames through compressed hidden states during tracking, thus releasing computational costs of handling temporal cues, even the initial target template.
    }
    \label{fig:architecture}
    \vspace{-3mm}
\end{figure*}

\subsection{SMTrack}
We design SASM blocks and present a neat temporal modeling paradigm for visual tracking, as shown in Figure~\ref{fig:architecture}.

\noindent
\textbf{SASM Block.}
In natural language processing, the original selective state space model block~\cite{Mamba} employs unidirectional scanning to process causal 1D sequences, enabling each element to interact with any of the previously scanned samples through a compressed hidden state.
In computer vision, many methods~\cite{Vim,VMamba,VideoMamba} introduce SSM via multi-directional scanning by flipping or transposing to process 2D images for spatial-aware feature learning.
Recently, MambaVT~\cite{MambaVT} integrates SSM to introduce temporal cues for RGBT tracking via bidirectional scanning.
By implementing a full flip of the sequence, target templates need to be scanned with the future information ($e.g.$ search region) in MambaVT, thus necessitating repeated scanning of templates for each frame during tracking.
Unlike previous methods, we seek to not only maintain the causality between frames for efficient temporal modeling but also learn spatial-aware features for image understanding in SASM blocks.
Thus, we design a temporal causal flipping method for temporal causal scanning as shown in Figure~\ref{fig:architecture}(a) (different colors represent different frames and different shades represent different image patches).
We only flip the sequence intra-frame so the temporal causality across frames can be maintained.
As shown in Figure~\ref{fig:architecture}(a), given the input~feature $F\in \mathbb{R}^{B\times L\times D}$, the SASM block can be formulated as,
\begin{gather}
    F_{x} = {\rm Linear}({\rm Norm}(F)), F_{z} = {\rm Linear}({\rm Norm}(F)), \\
    F_{f}^\prime = \sigma({\rm Conv1D}_f(F_{x})), \\
    F_{b}^\prime = \sigma({\rm Conv1D}_b({\rm TCFlip}(F_{x}))), \\
    F_{f},h_{last}^f = {\rm SASM}(F_{f}^\prime,h_{init}^f),\\
    F_{b},h_{last}^b = {\rm SASM}(F_{b}^\prime,h_{init}^b),\\
    F_{out} = {\rm Linear}(F_{f} \odot \sigma(F_{z}) + F_{b} \odot \sigma(F_{z})),
\end{gather}
where $\rm Norm(\cdot)$ is RMS normalization~\cite{RMSNorm}, $\sigma(\cdot)$ is SiLU function, $h_{init}^{f}, h_{init}^{b}\in \mathbb{R}^{B\times D\times N}$ are learnable initial hidden states of forward and backward scanning, $h_{last}^{f}, h_{last}^{b}\in \mathbb{R}^{B\times D\times N}$ are last hidden states of forward and backward scanning, $\rm TCFlip(\cdot)$ is temporal causal flipping.
For simplicity, SASMBlock can be expressed as,
\begin{gather}
    F_{out},h_{last}^f,h_{last}^b={\rm SASMBlock}(F,h_{init}^{f},h_{init}^{b}).
\end{gather}

\begin{algorithm*}[t]
\caption{Tracking pipeline of SMTrack.}
\small
\label{alg:algo}
\begin{algorithmic}[1]
\Require
    Frame sequence $\{I_i\}_{i=1}^N$, initialization bounding box $\hat{b}_1=b_1$, update interval $T$, memory of hidden states $\mathcal{M}=\emptyset$ and initial hidden states $\{h_{init}^{k,f}\}_{k=1}^{N_{SASM}},\{h_{init}^{k,b}\}_{k=1}^{N_{SASM}}$;
\Ensure
    Tracking results $\{\hat{b}_i\}_{i=2}^N$;
\State Crop initial template: $I_t \leftarrow {\rm CropTempate}(I_1,\hat{b}_1)$;
\State Patch embedding: $F_t^0 = {\rm PatchEmbed}(I_t)$;
\State Initialize template hidden states: $F_t^{N_{SASM}},\{h_{t,last}^{k,f}\}_{k=1}^{N_{SASM}},\{h_{t,last}^{k,b}\}_{k=1}^{N_{SASM}}\leftarrow{\rm SASMBlocks}(F_t^0,\{h_{init}^{k,f}\}_{k=1}^{N_{SASM}},\{h_{init}^{k,b}\}_{k=1}^{N_{SASM}})$;
\State Update memory of hidden states: $\mathcal{M}\leftarrow {\rm Append}(\mathcal{M},\{h_{t,last}^{k,f}\},\{h_{t,last}^{k,b}\})$;
\For {$i = 2,...,N$}
    \State Crop search region: $I_s \leftarrow {\rm CropSR}(I_i,\hat{b}_i)$;
    \State Patch embedding: $F_s^0 \leftarrow {\rm PatchEmbed}(I_s)$;
    \State Interact with hidden states: $F_s^{N_{SASM}},\{h_{s,last}^{k,f}\}_{k=1}^{N_{SASM}},\{h_{s,last}^{k,b}\}_{k=1}^{N_{SASM}}\leftarrow{\rm SASMBlocks}(F_s^0,\{h_{t,last}^{k,f}\}_{k=1}^{N_{SASM}},\{h_{t,last}^{k,b}\}_{k=1}^{N_{SASM}})$;
    \State Predict tracking box: $\hat{b}_i \leftarrow {\rm BoxHead}(F_s^{N_{SASM}})$;
    \If {$i$ mod $T=0$}
        \State Crop new template: $I_{nt} \leftarrow {\rm CropTempate}(I_i,\hat{b}_i)$;
        \State Patch embedding: $F_{nt}^0 = {\rm PatchEmbed}(I_{nt})$;
        \State Update template hidden states: 
        \Statex \qquad\qquad\qquad $F_{nt}^{N_{SASM}},\{h_{t,last}^{k,f}\}_{k=1}^{N_{SASM}},\{h_{t,last}^{k,b}\}_{k=1}^{N_{SASM}}\leftarrow{\rm SASMBlocks}(F_{nt}^0,\{h_{t,last}^{k,f}\}_{k=1}^{N_{SASM}},\{h_{t,last}^{k,b}\}_{k=1}^{N_{SASM}})$;
        \State Update memory of hidden states: $\mathcal{M}\leftarrow {\rm Append}(\mathcal{M},\{h_{t,last}^{k,f}\},\{h_{t,last}^{k,b}\})$;
        \State Sample and average hidden states in memory: $\{h_{t,last}^{k,f}\}_{k=1}^{N_{SASM}},\{h_{t,last}^{k,b}\}_{k=1}^{N_{SASM}} = {\rm SampleAvg}(\mathcal{M})$
    \EndIf
\EndFor
\end{algorithmic}
\end{algorithm*}

\noindent
\textbf{Training with Temporal Cues.}
Previous CNN-based or Transformer-based trackers commonly need numerous computational costs~\cite{cui2022mixformer,yan2021learning,APMT} or complex modules and pipelines~\cite{ARTrackV2,HIPTrack,AQATrack} to introduce long-range temporal cues for training.
Differently, our SMTrack can introduce more target templates with linear complexity to integrate long-range temporal cues for training without needing extra designing.
As shown in Figure~\ref{fig:architecture}(a), the target templates and search region are cropped multi-times the target size.
Given target templates $I_t\in\mathbb{R}^{N_t\times 3\times H_t\times W_t}$ and search region $I_s\in\mathbb{R}^{3\times H_s\times W_s}$, we split and map them into patch embeddings $F_t\in\mathbb{R}^{N_t\times L_t\times D}$ and $F_s\in\mathbb{R}^{L_s\times D}$ via linear projections.
Here, $N_t$ is the number of templates used for model training, $L_t=H_t W_t / p^2,L_s=H_s W_s / p^2$ are the patch number of the target template and search region, $p$ is the resolution of each patch.
Then, we add positional embeddings $E_t\in\mathbb{R}^{L_t\times D},E_s\in\mathbb{R}^{L_s\times D}$ for spatial-aware feature learning, add temporal embeddings $E_a\in\mathbb{R}^{N_t\times D}$ for temporal-aware feature learning and add target embeddings $E_b\in\mathbb{R}^{N_t\times L_t\times D}$ for target-aware feature learning,
\begin{gather}
    \hat{E}_t={\rm BoardCast}(E_t),\hat{E}_a={\rm BoardCast}(E_a)\\
    F_t^0 = F_t + \hat{E}_t + \hat{E}_a + E_b,\\
    F_s^0 = F_s + E_s,
\end{gather}
where $E_t,E_a,E_b$ are learnable parameters, target embeddings $E_b$ are generated by assigning positions in the target region with a parameter vector and assigning positions out of the target region with another parameter vector, temporal embeddings $E_a$ are used to indicate ordinal positions of templates.
We broadcast $E_t$ and $E_a$ in temporal and spatial dimensions respectively to align with the shape of $N_t \times L_t \times D$.
The features of the target template and search region are concatenated together $[F_t^0;F_s^0]$ and then fed into $N_{SASM}$ SASM blocks to build global interactions between templates and search region,
%
% \begin{small}
\begin{gather}
    [F_t^{i+1};F_s^{i+1}],h^{i,f}_{last},h^{i,b}_{last} = {\rm SASMBlock}^i([F_t^{i};F_s^{i}],h^{i,f}_{init},h^{i,b}_{init}),
\end{gather}
% \end{small}

\noindent
We take the enhanced features of search region $F_s^{N_{SASM}}$ as input of the box head to regress the bounding box.

Inspired by OSTrack~\cite{OSTrack}, we utilize a three-branch fully-convolutional network to regress a classification score map $\hat{\mathbf{C}}\in (0,1)^{\frac{H_x}{p}\times\frac{W_x}{p}}$, an offset map $\hat{\mathbf{O}}\in [0,1)^{2\times\frac{H_x}{p}\times\frac{W_x}{p}}$ and a normalized box size map $\hat{\mathbf{S}}\in (0,1)^{2\times\frac{H_x}{p}\times\frac{W_x}{p}}$ in the box head.
Given the position $(x_c,y_c)={\rm argmax}_{(x,y)}\hat{\mathbf{C}}(x,y)$, the bounding box of target $\hat{b}=(\hat{x},\hat{y},\hat{w},\hat{h})$ is formulated as,
% \begin{footnotesize} 
\begin{gather}
    (\hat{x},\hat{y}) = \Big(\big(x_c+\hat{\mathbf{O}}(0,x_c,y_c)\big)\cdot p,
                             \big(y_c+\hat{\mathbf{O}}(1,x_c,y_c)\big)\cdot p\Big),\\
    (\hat{w},\hat{h}) = \big(\hat{\mathbf{S}}(0,x_c,y_c)\cdot H_x, \hat{\mathbf{S}}(1,x_c,y_c)\cdot W_x\big).
\end{gather}
% \end{footnotesize}

%
We generate the groundtruth of the classification score map by a Gaussian kernel $\mathbf{C}(x,y)={\rm exp}\left(-\frac{(x-\widetilde{x})^2+(y-\widetilde{y})^2}{2\sigma^2}\right)$, where $(\widetilde{x}, \widetilde{y})$ is the groundtruth center of target object,
$\sigma$ is an object size-adaptive standard deviation~\cite{law2018cornernet}.
Then, the weighted focal loss~\cite{law2018cornernet} $\mathcal{L}_{focal}(\cdot,\cdot)$ is utilized for classification.
The $l_1$ loss $\mathcal{L}_{1}(\cdot,\cdot)$ and the generalized IoU loss~\cite{GIoU} $\mathcal{L}_{giou}(\cdot,\cdot)$ are utilized for bounding box regression.
Formally,
\begin{gather}
    \mathcal{L} = \mathcal{L}_{focal}(\mathbf{C},\hat{\mathbf{C}}) + \lambda_{1}\mathcal{L}_{1}(b,\hat{b}) + \lambda_{giou}\mathcal{L}_{giou}(b,\hat{b}).
\end{gather}

\noindent
\textbf{Tracking with Previous States.}
We maintain the temporal causality between frames in SASM blocks.
Thus, the search region can freely build interactions with scanned templates via hidden states without the need for repeated scanning of templates during tracking, as shown in Figure~\ref{fig:architecture}(b).
% 控制状态的更新来平衡效率与性能
Given the patch embeddings of the initial target template $F_t^0$, we can acquire the last hidden states $h_{t,last}^{i,f},h_{t,last}^{i,b}$ of the template in each SASM block. Formally,
%
% \begin{small}
\begin{gather}
    F_t^{i+1},h_{t,last}^{i,f},h_{t,last}^{i,b}={\rm SASMBlock}^i(F_t^i,h_{init}^{i,f},h_{init}^{i,b}).
\end{gather}
% \end{small}

\noindent
Then, search region embeddings $F_s^0$ can build interactions with the template via the scanned hidden states $h_{t,last}^{i,f},h_{t,last}^{i,b}$, which can be formulated as,
%
% \begin{small}
\begin{gather}
    F_s^{i+1},h_{s,last}^{i,f},h_{s,last}^{i,b}={\rm SASMBlock}^i(F_s^i,h_{t,last}^{i,f},h_{t,last}^{i,b}).
\end{gather}
% \end{small}

\noindent
Then, the features of the search region are fed into the box head to localize the target. 
Further, we can update $h_{t,last}^{i,f},h_{t,last}^{i,b}$ via new target templates of tracked frames $F_{nt}^0$ to introduce temporal cues,
%
% \begin{small}
\begin{gather}
    F_{nt}^{i+1},h_{t,last}^{i,f},h_{t,last}^{i,b}={\rm SASMBlock}^i(F_{nt}^i,h_{t,last}^{i,f},h_{t,last}^{i,b}).
\end{gather}
% \end{small}

\noindent
% 不需要复杂的核优化器或者跨帧关联的设计，这简化了跟踪的流程。
% 同时可以建立全局非线性交互，使能更鲁棒的跟踪能力。
Thanks to our effective designing, SMTrack can build global interactions between the target template and search region via compressed hidden states.
% SMTrack不需要在每一帧与模板进行密集交互，而是隐状态的传递这降低了计算代价。
% SMTrack不需要设计复杂的模块与训练策略来学习压缩的时序信息，这简化了网络设计。
Meanwhile, SMTrack can easily introduce new temporal cues via hidden state updating without needing extra module designing or numerous computational costs, which provides a neat network structure and tracking pipeline for temporal modeling.

%-------------------------------------------------------------------------

\section{Experiments}
\label{sec:exp}

\subsection{Implementation Details}
\label{sec:imp_details}
Our trackers are implemented using Python 3.8 and Pytorch 2.4.1. The SMTrack is trained on a server with eight 24GB NVIDIA RTX 3090 GPUs and AMD EPYC 7713 64-Core Processor @ 2GHz with 503 GB RAM.

%
% \begin{table}[h]
%     \centering
%     \caption{Input resolutions of SMTrack variants}
%     \label{tab:variant}
%     \resizebox{0.8\linewidth}{!}{
%     \begin{tabular}{c|c|c}
%         \hline
%         \multirow{2}{*}{variants} & \multicolumn{2}{c}{resolutions} \\\cline{2-3}
%         & target template & search region \\\hline
%         SMTrack-S256 & 256$\times$256 & 256$\times$256 \\
%         SMTrack-M256 & 128$\times$128 & 256$\times$256 \\
%         SMTrack-M384 & 192$\times$192 & 384$\times$384 \\\hline
%     \end{tabular}
%     }
% \end{table}

\noindent
\textbf{Network details}. 
To demonstrate the scalability of SMTrack, we present three variants of SMTrack with different input resolutions, including SMTrack-S256, SMTrack-M256 and SMTrack-M384.
The details are shown in Table~\ref{tab:variant}.
We crop target templates and search regions by $4^2$ and $4^2$ times the target bounding box area for SMTrack-S256.
We crop target templates and search regions by $2^2$ and $4^2$ times the target bounding box area for SMTrack-M256/384.
%
% These images are split into 16$\times$16 patches as the input of SMTrack.
%
We stack 24 SASM blocks with feature channel $D=384$ for SMTrack-S256, and 32 SASM blocks with feature channel $D=576$ for SMTrack-M256/384.
The number of hidden states is set as $N=16$.
SMTrack-S256 and SMTrack-M256/384 parameters are initialized with pretrained parameters of VideoMamba-S and VideoMamba-M~\cite{VideoMamba} respectively.

\begin{table}[t]
    \centering
    \caption{Hyperparamete setting}
    \label{tab:variant}
    \resizebox{0.8\linewidth}{!}{
    \begin{tabular}{c|cccc}\hline
         & $(H_t,W_t)$ & $(H_s,W_s)$ & $N_{SASM}$ & $D$ \\\hline
        SMTrack-S256 & (256,256) & (256,256) & 24 & 384 \\
        SMTrack-M256 & (128,128) & (256,256) & 32 & 576 \\
        SMTrack-M384 & (192,192) & (384,384) & 32 & 576 \\\hline
        \multicolumn{5}{c}{\textit{Common:} $N=16$, $\lambda_1=2$, $\lambda_{giou}=5$, $|\mathcal{M}|=50$}\\
        \hline
    \end{tabular}}
\end{table}

% \usepackage{multirow}
\begin{table*}[!t]
\begin{center}
\caption{Comparison with state-of-the-art visual trackers on GOT10k, TrackingNet, LaSOT, LaSOT$_{ext}$, UAV and NFS. The best two results are shown in \textbf{\textcolor{red}{red}} and \textbf{\textcolor{blue}{blue}} for each category of network architecture.
}
\label{tab:mainresults}
\resizebox{0.90\linewidth}{!}{
\begin{tabular}{l|ccc|ccc|ccc|ccc|c|c} 
\toprule
\multicolumn{1}{c|}{\multirow{2}{*}{Method}} & \multicolumn{3}{c|}{GOT-10k} & \multicolumn{3}{c|}{Trackingnet} & \multicolumn{3}{c|}{LaSOT} & \multicolumn{3}{c|}{LaSOT$_{ext}$} & UAV  & NFS   \\ 
\cline{2-15}
\multicolumn{1}{c|}{}                        & AO   & SR$_{0.5}$ & SR$_{0.75}$        & AUC  & P    & P$_{norm}$              & AUC  & P    & P$_{norm}$        & AUC  & P    & P$_{norm}$           & AUC  & AUC   \\ 
\hline
\multicolumn{15}{c}{SSM-based Tracker}                                                                                                                                                     \\ 
\hline
SMTrack-M384                                & \textbf{\textcolor{red}{74.7}}  & \textbf{\textcolor{blue}{83.4}}   & \textbf{\textcolor{red}{72.1}}                & \textbf{\textcolor{red}{85.2}}  & \textbf{\textcolor{red}{85.0}}  & \textbf{\textcolor{red}{90.0}}           & \textbf{\textcolor{red}{71.9}}  & \textbf{\textcolor{red}{79.2}}  & \textbf{\textcolor{red}{81.4}}          & \textbf{\textcolor{red}{51.6}}  & \textbf{\textcolor{red}{59.0}}  & \textbf{\textcolor{red}{62.1}}             & \textbf{\textcolor{red}{70.8}}  & \textbf{\textcolor{red}{69.3}}   \\
SMTrack-M256                                & \textbf{\textcolor{blue}{74.5}}  & \textbf{\textcolor{red}{84.6}}   & \textbf{\textcolor{blue}{71.0}}           & \textbf{\textcolor{blue}{84.2}}  & \textbf{\textcolor{blue}{82.9}}  & \textbf{\textcolor{blue}{89.2}}                & \textbf{\textcolor{blue}{70.1}}  & \textbf{\textcolor{blue}{76.9}}  & \textbf{\textcolor{blue}{80.1}}          & \textbf{\textcolor{blue}{48.9}}  & \textbf{\textcolor{blue}{55.1}}  & 59.7             & \textbf{\textcolor{blue}{69.8}}  & \textbf{\textcolor{blue}{67.2}}   \\
SMTrack-S256                                & 71.7  & 82.3   & 64.9           & 82.7  & 80.8  & 88.0                & 69.0  & 75.3  & 79.8          & 48.4  & 54.7  & \textbf{\textcolor{blue}{59.8}}             & 68.8  & 66.0   \\ 
\hline
\multicolumn{15}{c}{Transformer-based Tracker}                                                                                                                                             \\ 
\hline
% HIPTrack & 77.4 & 88.0 & 74.5 & 84.5 & 89.1 & 83.8 & 72.7 & 82.9 & 79.5 & 53.0 & 64.3 & 60.6 & 70.5 & 68.1 &
ROMTrack-384~\cite{ROMTrack}                                 & 74.2 & \textbf{\textcolor{blue}{84.3}}  & \textbf{\textcolor{red}{72.4}}     &\textbf{\textcolor{red}{84.1}} & \textbf{\textcolor{red}{83.7}} & \textbf{\textcolor{red}{89.0}}   & \textbf{\textcolor{blue}{71.4}} & \textbf{\textcolor{red}{78.2}} & \textbf{\textcolor{red}{81.4}}     & \textbf{\textcolor{red}{51.3}} &\textbf{\textcolor{red}{58.6}}  &\textbf{\textcolor{red}{62.4}}     & -    & \textbf{\textcolor{red}{68.8}}  \\
ROMTrack~\cite{ROMTrack}                                     & 72.9 & 82.9  & 70.2          & 83.6 & 82.7 & 88.4               & 69.3 & 75.6 & 78.8         & 48.9 & 55.0 & 59.3            & -    & \textbf{\textcolor{blue}{68.0}}  \\
VideoTrack~\cite{VideoTrack}                                   & 72.9 & 81.9  & 69.8          & 83.8 & 83.1 & 88.7               & 70.2 & 76.4 & -            & -    & -    & -               & 69.7 & -     \\
SeqTrack-B384~\cite{seqtrack}                                & \textbf{\textcolor{blue}{74.5}} & 84.3  & 71.4          & \textbf{\textcolor{blue}{83.9}} & \textbf{\textcolor{blue}{83.6}} & \textbf{\textcolor{blue}{88.8}}     & \textbf{\textcolor{red}{71.5}} & \textbf{\textcolor{blue}{77.8}} & \textbf{\textcolor{blue}{81.1}}         & \textbf{\textcolor{blue}{50.5}} & 57.5 & \textbf{\textcolor{blue}{61.6}}            & 68.6 & 66.7  \\
SeqTrack-B256~\cite{seqtrack}                                & \textbf{\textcolor{red}{74.7}} & \textbf{\textcolor{red}{84.7}}  & \textbf{\textcolor{blue}{71.8}}          & 83.3 & 82.2 & 88.3               & 69.9 & 76.3 & 79.7         & 49.5 & 56.3 & 60.8            & 69.2 & 67.6  \\
OSTrack-384~\cite{OSTrack}                                  & 73.7 & 83.2  & 70.8          & 83.9 & 83.2 & 88.5               & 71.1 & 77.6 & 81.1         & 50.5 & \textbf{\textcolor{blue}{57.6}} & 61.3            &\textbf{\textcolor{red}{70.7}} & 66.5  \\
OSTrack-256~\cite{OSTrack}                                  & 71.0 & 80.4  & 68.2          & 83.1 & 82.0 & 87.8               & 69.1 & 75.2 & 78.7         & 47.4 & 53.3 & 57.3            & 68.3 & 64.7  \\
MixViT~\cite{mixvit}         & 72.7 & 82.3 & 70.8       & 83.5 & 82.0 & 88.3        & 69.6 & 75.9 & 69.9       & - & - & - & 68.1 & - \\
MixFormer~\cite{cui2022mixformer}                                & 70.7 & 80.0  & 67.8          & 83.1 & 81.6 & 88.1               & 69.2 & 74.7 & 78.7         & -    & -    & -               & \textbf{\textcolor{blue}{70.4}} & -     \\
STMTrack~\cite{fu2021stmtrack}                                     & 64.2 & 73.7  & 57.5          & 80.3 & 76.7 & 85.1               & 60.6 & 63.3 & 69.3         & -    & -    & -               & 64.7 & -     \\
TrDiMP~\cite{TMT}                                       & 68.8 & 80.5  & 59.7          & 78.4 & 73.1 & 83.3               & 63.9 & 61.4 & -            & -    & -    & -               & 67.5 & 66.5  \\
STARK~\cite{yan2021learning}                                        & 68.8 & 78.1  & 64.1          & 82.0 & -    & 86.9               & 67.1 & -    & 77.0         & -    & -    & -               & 69.1 & 66.2  \\
TransT~\cite{TransT}                                       & 72.3 & 82.4  & 68.2          & 81.4 & 80.3 & 86.7               & 64.9 & 69.0 & 73.8         & -    & -    & -               & 69.1 & 65.7  \\ 
\hline
\multicolumn{15}{c}{CNN-based Tracker}                                                                                                                                                     \\ 
\hline
KeepTrack~\cite{mayer2021learning}                                    & -    & -     & -             & -    & -    & -                  & \textbf{\textcolor{red}{67.1}} & \textbf{\textcolor{red}{70.2}} & \textbf{\textcolor{red}{77.2}}         & \textbf{\textcolor{red}{48.2}} & -    & -               & \textbf{\textcolor{red}{69.7}} & \textbf{\textcolor{red}{66.4}}  \\
SiamRCNN~\cite{SiamRCNN}                                     & \textbf{\textcolor{red}{64.9}} & 72.8  & \textbf{\textcolor{red}{59.7}}      &\textbf{\textcolor{red}{81.2}}  & \textbf{\textcolor{red}{80.0}} & \textbf{\textcolor{red}{85.4}}    & \textbf{\textcolor{blue}{64.8}} & -    & \textbf{\textcolor{blue}{72.2}}         & -    & -    & -               & 64.9 & \textbf{\textcolor{blue}{63.9}}  \\
KYS~\cite{KYS}                                          & \textbf{\textcolor{blue}{63.6}} &\textbf{\textcolor{red}{75.1}}   & 51.5          & 74.0 & 68.8 & 80.0         & -    & -    & -                  & -    & -    & -               & -    & 63.5  \\
PrDiMP~\cite{PrDiMP}                                       & 63.4 & \textbf{\textcolor{blue}{73.8}} & \textbf{\textcolor{blue}{54.3}}         & \textbf{\textcolor{blue}{75.8}} & \textbf{\textcolor{blue}{70.4}} & \textbf{\textcolor{blue}{81.6}}               & 59.8 & -    & -            & -    & -    & -               & \textbf{\textcolor{blue}{68.0}} & 63.5  \\
DiMP~\cite{DiMP}                                         & 61.1 & 71.7  & 49.2          & 74.0 & 68.7 & 80.1               & 56.9 & \textbf{\textcolor{blue}{56.7}} & 65.0         & \textbf{\textcolor{blue}{39.2}} & \textbf{\textcolor{red}{45.1}} & \textbf{\textcolor{red}{47.6}}            & 65.3 & 61.9  \\
% SiamFC++                                     & 59.5 & 69.5  & 47.9          & 75.4 & 70.5 & 80.0               & 54.4 & -    & -            & -    & -    & -               & -    & -     \\
Ocean~\cite{Ocean}                                        & 61.1 & 72.1  & 47.3          & -    & -    & -                  & 56.0 & 56.6 & 65.1         & -    & -    & -               & -    & -     \\
SiamBAN~\cite{SiamBAN}                                      & -    & -     & -             & -    & -    & -                  & 51.4 & -    & 59.8         & -    & -    & -               & 63.1 & 59.4  \\
SiamCAR~\cite{SiamCAR}                                      & 56.9 & 67.0  & 41.5          & -    & -    & -                  & 50.7 & 51.0 & 60.0         & -    & -    & -               & 61.4 & -     \\
UpdateNet~\cite{UpdateNet}                                    & -    & -     & -             & 67.7 & 62.5 & 75.2               & 47.5 & -    & 56.0         & -    & -    & -               & -    & -     \\
ATOM~\cite{ATOM}                                         & 55.6 & 63.4  & 40.2          & 70.3 & 64.8 & 77.1               & 51.5 & 50.5 & 57.6         & 37.6 & \textbf{\textcolor{blue}{43.0}} &\textbf{\textcolor{blue}{45.9}}            & 65.0 & 59.0  \\
SiamRPN++~\cite{SiamRPNplusplus}                                    & 51.7 & 61.6  & 32.5          & 73.3 & 69.4 & 80.0               & 49.6 & 49.1 & 56.9         & 34.0 & 39.6 & 41.6            & 61.3 & -     \\
% ECO                                          & 31.6 & 30.9  & 11.1          & 55.4 & 49.2 & 61.8               & 32.4 & 30.1 & 33.8         & 22.0 & 24.0 & 25.2            & 53.7 & -     \\
% SiamFC                                       & 34.8 & 35.3  & 9.8           & 57.1 & 53.3 & 66.3               & 33.6 & 33.9 & 42.0         & 23.0 & 26.9 & 31.1            & 46.8 & -     \\
\bottomrule
\end{tabular}
}
\end{center}
\end{table*}

\noindent
\textbf{Training details}. 
We train our model on the training splits of COCO~\cite{COCO}, LaSOT~\cite{LaSOT}, GOT-10K~\cite{GOT10K}, and TrackingNet~\cite{trackingnet}, which is a general setting for model training~\cite{yan2021learning,OSTrack} in visual tracking.
Common data augmentation techniques, including translation, horizontal flipping, and brightness jittering, are employed to enhance model robustness~\cite{OSTrack}.
The minimal training data unit for SMTrack consists of four target templates and one search region.
$\lambda_1=2$ and $\lambda_{giou}=5$ in the training objective.
Moreover, we train our SMTrack by the AdamW optimizer~\cite{AdamW} with the weight decay $10^{-4}$.
The learning rates start from $4\times10^{-5}$ for SASM blocks and $4\times10^{-4}$ for the box head, then gradually reduce to 0 with cosine annealing~\cite{CosLR}.
Our model is trained with a batch size of 96 and iterates 300 epochs with $6\times10^5$ samples per epoch.
In order to evaluate the performance of SMTrack on GOT-10k~\cite{GOT10K} benchmark, we follow the one-shot tracking rule of GOT-10k and retrain our model for 100 epochs with $6\times 10^5$ samples per epoch on the GOT-10k train split.
The other training settings are consistent with the models trained on entire datasets.

\noindent
\textbf{Inference details}. 
We scan the initial template once and save the last hidden states of each SASM block.
Then, the search region of each frame builds interactions with the template via saved hidden states.
%
% Also, we update the hidden states through new templates of tracked frames every 20 frames.
%
Thanks to the linear form interactions between image features and hidden states in Equation~\ref{equ:4}, we can easily establish the interactions between the search region and multiple target templates via the sum of their hidden states without needing extra computational costs.
So, we sample multiple hidden states of previous templates to build multi-frame longer-range dependencies with the search region so that more temporal cues can be integrated for target discrimination and avoid the accumulation of errors.
Details are as follows.
We crop a new target template every 20 frames using the tracking results and scan the target template to obtain the last hidden states of each SASM block, which are saved into memory $\mathcal{M}$.
During tracking, we evenly sample $N_h$ hidden states $\{\hat{h}^{(j)}\}_{j=1}^{N_h}$ from memory $\mathcal{M}$. 
Here $\hat{h}^{(j)}$ denotes the $j^{th}$ sampled hidden state, which contains the last hidden states of forward and backward scanning in each SASM block $\left\{h_{t,last}^{i,f},h_{t,last}^{i,b}\right\}_{i=1}^{N_{SASM}}$.
The index of hidden states in memory is determined by the formula below,
\begin{gather}
    \left\{\lfloor\frac{i\cdot(|\mathcal{M}|-1)}{N_h-1}\rfloor+1\right\}_{i=0}^{N_h-1},
\end{gather}
Here, $|\mathcal{M}|$ denotes the memory size, $N_h$ is set as 10.
After that, we take the average of the sampled hidden states as the initial states of each SASM block for the search region.
The max size of the memory is set as 50.
When the maximum memory size is exceeded, we discard the hidden states whose frame indexes are closest to each other to ensure the temporal diversity in memory.
This design allows the search region to interact with target templates of almost all time ranges with little computational costs.
The tracking pipeline of SMTrack is show in Algorithm~\ref{alg:algo}.
More details will be discussed in Section~\ref{sec:ablation}.

\subsection{State-of-the-art Comparisons}

Here, we are the first tracker with a pure SSM feature extractor for RGB tracking.
Notably, MambaTV~\cite{MambaVT} is designed for RGBT tracking. 
Thus, it is evaluated on completely different benchmarks.
Nevertheless, we performed ablation experiments using MambaVT-like architecture in Section~\ref{sec:ablation}.

\noindent
\textbf{Metrics}.
The average overlap $AO$~\cite{GOT10K} denotes the average of overlaps between all groundtruth and estimated bounding boxes.
The success rate $SR$~\cite{GOT10K} measures the percentage of successfully tracked frames where the overlaps exceed a threshold ($e.g$., 0.5, 0.75).
The area under the curve $AUC$~\cite{trackingnet} denotes the average of the success rates corresponding to the sampled overlap thresholds.
The precision $P$~\cite{trackingnet} is measured as the distance in pixels between the centers of the groundtruth and the tracker bounding box with a threshold of 20 pixels.
The normalized precision $P_{norm}$~\cite{trackingnet} further normalize the precision over the size of the groundtruth bounding box.

\noindent
\textbf{GOT-10k}.
GOT-10k~\cite{GOT10K} is a large-scale benchmark dataset comprising 10,000 sequences for training and 180 sequences for testing, with a protocol that mandates training trackers exclusively on the training split of GOT-10k.
As shown in Table~\ref{tab:mainresults}, our SMTrack-M256 outperforms previous temporal modeling trackers, VideoTrack and ROMTrack by 1.6\% AO.
Further, SMTrack-M384 achieves the best AO score of 74.7\%.
This proves the superiority of our temporal modeling paradigm.

\noindent
\textbf{TrackingNet}.
TrackingNet~\cite{trackingnet} is a large-scale dataset consisting of over 30,000 videos sampled from YouTube, with a testing set of 511 videos.
Our SMTrack-M256 performs better than VideoTrack and ROMTrack, and SMTrack-M384 outperforms ROMTrack-384 by 1.1\% AUC.

\noindent
\textbf{LaSOT}.
LaSOT~\cite{LaSOT} is a large-scale benchmark for long-term tracking, comprising a total of 1,400 videos, with 280 videos designated for the testing set.
As reported in Table~\ref{tab:mainresults}, SMTrack-M384 achieves the best performance of 71.9\% AUC, showing the superiority of our temporal modeling paradigm in long-term tracking.

\noindent
\textbf{LaSOT$_{ext}$}.
LaSOT$_{ext}$~\cite{LaSOT_ext} is an extended version of the LaSOT dataset, containing 150 videos spanning 15 additional categories.
As shown in Table~\ref{tab:mainresults}, SMTrack-M384 performs better than ROMTrack-384 with the AUC score of 51.6\%.

\noindent
\textbf{UAV}.
UAV123~\cite{UAV} dataset consists of 123 aerial videos captured by an unmanned aerial vehicle (UAV).
Our SMTrack-M384 obtains the best AUC score of 70.8\%.

\noindent
\textbf{NFS}.
The Need for Speed (NFS)~\cite{NFS} dataset comprises 100 challenging videos captured using a high-frame-rate camera.
The results in Table~\ref{tab:mainresults} show that SMTrack-M384 achieves the best performance of 69.3\% AUC.

\begin{table}[t]
    \centering
    \caption{Comparison of consecutive temporal modeling trackers.}
    \label{tab:efficiency}
    \resizebox{\linewidth}{!}{
    \begin{tabular}{c|ccccc}\hline
        Method & HIPTrack~\cite{HIPTrack} & AQATrack~\cite{AQATrack} & EVPTrack~\cite{EVPTrack} & ODTrack~\cite{ODTrack} & SMTrack-M384 \\ \hline
        Trackingnet AUC $\uparrow$ & 84.5 & 84.8 & 84.4 & 85.1 & \textcolor{red}{\textbf{85.2}} \\ 
        UAV AUC         $\uparrow$ & 70.5 & \textcolor{red}{\textbf{71.2}} & 70.9 & -    & 70.8 \\ \hline
        GFlops        $\downarrow$ & 66.9 & 62.1 & 78.9 & 86.2 & \textcolor{red}{\textbf{48.7}} \\ \hline
    \end{tabular}
    }
\end{table}

\noindent
\textbf{Compared with CNN-based Temporal Modeling Trackers}.
%
% As shown in Table~II, our SMTrack has significant performance advantages over previous CNN-based temporal modeling trackers.
%
Due to the limitations of global modeling in CNNs, popular CNN-based trackers perform temporal modeling through complex global detection and association mechanisms, which incur substantial computational costs.
For instance, KeepTrack~\cite{mayer2021learning} requires 1334.8 GFLOPs, whereas our SMTrack-M256 achieves only 21.7 GFLOPs while consistently outperforming KeepTrack on diverse benchmarks, as shown in Table~\ref{tab:mainresults}.
%
% This is because SMTrack can build global interactions with long-range temporal cues and introduce temporal cues via hidden state updating and propagation with little computational costs.
%
% Thanks to the state propagation of the selective state-aware space model (SASM) and temporal causality scanning, 
This is because SMTrack can naturally integrate temporal cues and build global interactions via state updating and propagation without needing extra module designs, such as association networks, providing a neat network structure and tracking pipeline with less computational costs.

\noindent
\textbf{Compared with Transformer-based Trackers}.
% 8.1  22.7
As shown in Table~\ref{tab:mainresults}, our SMTrack-S256 (8.5 Gflops) achieves competitive performance against the popular Transformer-based tracker, OSTrack-256 (22.7 Gflops), with only \textbf{35.7\%} computational costs.
Meanwhile, our SMTrack-M384 (48.7 Gflops) outperforms the autoregressive Transformer-based tracker, SeqTrack-B384 (148 Gflops), with only \textbf{32.9\%} computational costs.
Our SMTrack has the same neat network structure as theirs, while SMTrack can achieve superior performance with significantly less computational costs.
This is because SMTrack can naturally introduce temporal cues via hidden state updating and propagation with little computational costs.
%
% Therefore, SMTrack can achieve superior performance with less computational costs.
%
These results demonstrate the advantages of our SMTrack compared with previous Transformer-based trackers.

\noindent
\textbf{Compared with Transformer-based Temporal Modeling Trackers}.
We compare SMTrack with recent consecutive temporal modeling trackers in Table~\ref{tab:efficiency}.
Our SMTrack-M384 achieves competitive performance against these trackers with less computational cost.
Notably, these trackers commonly design customized modules to propagate temporal cues, such as temporal tokens~\cite{ODTrack}, visual prompts~\cite{HIPTrack,EVPTrack} and autoregressive queries~\cite{AQATrack}, leading to complex network structure.
Also, they desire to sample multiple search frames and predict them sequentially for temporal cue propagation during training, which complicates the training pipeline.
Unlike them, SMTrack builds interactions between search region and temporal cues directly via hidden state propagation and updating without needing extra module designing, providing a neat network structure and training pipeline for long-range temporal modeling.

\noindent
\textbf{Compared with Hybrid Temporal Modeling Trackers}.
Recent TemTrack~\cite{temtrack} employs a SSM-attention hybrid network after the Fast-iTPN~\cite{fast-itpn} backbone to model temporal appearance changes.
Our method achieves comparable performance (85.2\% AUC $v.s.$ 85.0\% AUC on TrackingNet) to TemTrack while requiring fewer computational costs (48.7 GFlops $v.s.$ 55.7 GFlops).
SMTrack presents a fundamentally different paradigm through its pure SSM-based architecture. 
SMTrack innovatively integrates feature extraction and temporal interaction within a streamlined SSM structure, leveraging the updating and propagation of hidden states to achieve temporal modeling with significantly reduced computational complexity.

% \noindent
% \textbf{Computational Efficiency}.
% %
% SMTrack-S256 (8.46GFlops) achieves competitive performance against OSTrack-256 (22.7GFlops) with only \textbf{37.3\%} computational costs.
% %
% SMTrack-M384 (48.7GFlops) outperforms SeqTrack-384 (148GFlops) with only \textbf{32.9\%} computational costs.
% %
% These results proves the superior of SMTrack against previous popular Transformer-based trackers.
% %
% As shown in Tabel~\ref{tab:efficiency}, SMTrack-M256 outperforms Transformer-based trackers of temporal modeling using no more than \textbf{58.2\%} computational costs.
% %
% Notably, SMTrack gets rid of extra module designing or complex training pipeline, which prove the superiority of our temporal modeling paradigm.

\subsection{Ablation Study}
\label{sec:ablation}
We take the ablation study of SMTrack to analyze its designs. The following experiments adopt SMTrack-M256. 

\begin{table}[t]
    \centering
    \caption{Analysis of different designs in SASM blocks on LaSOT.}
    \label{tab:ablation_SASM}
    \resizebox{\linewidth}{!}{
    \begin{tabular}{c|cccccc|c}
        \hline
        \# & $E_{t}$\&$E_{s}$ & $E_{a}$ & $E_{b}$ & Channel-wise $\Delta$ & State-wise $\Delta$ & Interaction & AUC $\uparrow$ \\\hline
        1 & - & - & - & \checkmark & - & - & 66.8 \\
        2 & \checkmark & - & - & \checkmark & - & - & 67.3 \\
        3 & \checkmark & \checkmark & - & \checkmark & - & - & 67.6 \\
        4 & \checkmark & \checkmark & \checkmark & \checkmark & - & - & 68.5 \\
        5 & \checkmark & \checkmark & \checkmark & - & \checkmark & - & 48.2 \\
        6 & \checkmark & \checkmark & \checkmark & \checkmark & \checkmark & - & 69.6 \\
        7 & \checkmark & \checkmark & \checkmark & \checkmark & \checkmark & \checkmark & \textbf{70.1} \\
        \hline
    \end{tabular}
    }
\end{table}

\begin{table}[t]
    \centering
    \caption{Different designs of state-wise parameters.}
    \label{tab:ablation_sw_p}
    \resizebox{0.9\linewidth}{!}{
    \begin{tabular}{c|ccccc}
        \hline
        Method  & $SoftPlus_1$ & $SoftPlus_2$ & $SoftPlus_3$ & $SoftPlus_4$ \\\hline
        AUC     & \textbf{70.1} & 69.8 & 68.7 & 65.2 \\
        \hline
    \end{tabular}
    }
\end{table}

\begin{table}[t]
    \centering
    \caption{Comparison of different temporal modeling paradigms on LaSOT.}
    \label{tab:ablation_temporal}
    \resizebox{1.0\linewidth}{!}{
    \begin{tabular}{c|ccccc}
        \hline
        Method  & $Trans_1$ & $Trans_3$ & $Trans_5$ & $BiSSM_5$ & SMTrack-M256 \\\hline
        % AO \uparrow     & 71.0 & 73.6 & 73.9 & 74.2 & \textbf{74.5} \\
        AUC $\uparrow$     & 68.5 & 69.5 & 69.7 & 69.8 & \textbf{70.1} \\
        GFlops $\downarrow$ & 30.9 & 43.7 & 57.1 & 46.4 & \textbf{21.7} \\
        \hline
    \end{tabular}
    }
\end{table}

\begin{table}[t]
    \centering
    \caption{Comparison of different numbers of hidden states on LaSOT.}
    \label{tab:hidden_state}
    \resizebox{0.7\linewidth}{!}{
    \begin{tabular}{c|ccccc}
        \hline
        $N$  & 4 & 8 & 12 & 16 & 24 \\\hline
        AUC $\uparrow$     & 63.2 & 67.1 & 68.8& 70.1 & \textbf{70.2} \\
        \hline
    \end{tabular}
    }
\end{table}

\begin{table}[t]
    \centering
    \caption{Performance of different update intervals on LaSOT.}
    \label{tab:ablation_updating}
    \resizebox{0.65\linewidth}{!}{
    \begin{tabular}{c|ccccc}
        \hline
        w/o TE  & 5 & 10 & 20 & 30 & 50 \\\hline
        AUC $\uparrow$       & 69.6 & 69.9 & \textbf{70.1} & 70.0 & 69.4 \\\hline
        w/ TE  & 5 & 10 & 20 & 30 & 50 \\\hline
        AUC $\uparrow$       & 70.0 & \textbf{70.2} & 70.1 & 69.8 & 69.0 \\
        \hline
    \end{tabular}
    }
\end{table}
\begin{table}[t]
    \centering
    \caption{Analysis of different propagation strategies on LaSOT.}
    \label{tab:ablation_propagation}
    \resizebox{1.0\linewidth}{!}{
    \begin{tabular}{c|cccccc}
        \hline
        Sample  & $Last_1$ & $Last_{5}$ & $Last_{10}$ & $Uni_{5}$ & $Uni_{10}$  & $Uni_{20}$ \\\hline
        AUC  $\uparrow$      & 68.5 & 69.2 & 69.3 & 69.8 & \textbf{70.1} & 70.0 \\
        \hline
    \end{tabular}
    }
\end{table}

\begin{table}[tp!]
    \centering
    \caption{Comparison of different memory size on LaSOT.}
    \label{tab:memory_size}
    \resizebox{0.6\linewidth}{!}{
    \begin{tabular}{c|ccccc}
        \hline
        $|\mathcal{M}|$  & 15 & 30 & 50 & 100  \\\hline
        AUC $\uparrow$     & 69.6 & 69.9 & 70.1& 70.1  \\
        \hline
    \end{tabular}
    }
\end{table}

% embeddings, State-wise, interaction, 
\noindent
\textbf{Effectiveness of the SASM block}.
As shown in Table~\ref{tab:ablation_SASM}, the positional embedding $E_t$,$E_s$, temporal embedding $E_a$ and target embedding $E_a$ are all vital for SMTrack to learn more effective features.
When we directly replace the channel-wise timescale parameters with state-wise timescale parameters, the performance drops dramatically.
This is because different channels encode distinct aspects of target-background information, with varying importance for discriminative feature learning.
Using only state-wise timescale parameters reduces the ability to assess the importance of channels and fails to take full advantage of the knowledge in the pretrained parameters.
Our SASM employs both channel-wise and state-wise timescale parameters $\Delta$, which enables hidden states to capture diverse temporal cues, facilitating the target discrimination for SMTrack, which brings 1.1\% AUC gains.
Also, the interactions between hidden states enable them to build dense dependencies between each other, which brings 0.5\% AUC gains.

\noindent
\textbf{Analysis of State-wise Parameter Designs}.
We try different ways to construct the state-wise parameters, as shown in Table~\ref{tab:ablation_sw_p}.
Given $\Delta_{state}\in\mathbb{R}^{B\times L \times N}$ and $\Delta_{channel}\in\mathbb{R}^{B\times L \times D}$, we broadcast $\Delta_{state}$ in $3^{rd}$ dimension and $\Delta_{channel}$ in $4^{th}$ dimension, making them have the same shape of $(B, L, D, N)$.
As shown in Table~\ref{tab:ablation_sw_p},
$SoftPlus_1$ means $\Delta=\sigma(\Delta_{state} + \Delta_{channel})$.
$SoftPlus_2$ means $\Delta=\sigma(\Delta_{state}) + \sigma(\Delta_{channel})$.
$SoftPlus_3$ means $\Delta={\rm sigmoid}(\Delta_{state})\cdot\sigma(\Delta_{channel})$.
$SoftPlus_4$ means $\Delta=\sigma(\Delta_{state})\cdot{\rm sigmoid}(\Delta_{channel})$.
Here, $\sigma$ means softplus operation.
As we can see, $SoftPlus_1$ and $SoftPlus_2$ are generally similar in performance.
However, $SoftPlus_3$ and $SoftPlus_4$ have significant performance degradation.
The underlying reason could be that the softplus operation is easier to be optimized from the gradient perspective and thus suitable as a gating mechanism for SASM, as Mamba~\cite{Mamba} does.

\noindent
\textbf{Effectiveness of the new Temporal Modeling Paradigm}.
We compare our SMTrack with other temporal modeling paradigms based on Transformer and bidirectional SSM (BiSSM).
Here, BiSSM shares the similar architecture with SMTrack but introduces two key distinctions. 
First, BiSSM employs a full flipping mechanism to enable bidirectional state propagation across frames, whereas SMTrack utilizes temporal causal flipping that restricts state propagation to unidirectional causality. 
Second, BiSSM explicitly maintains a memory bank of raw target templates (instead of hidden states as in SMTrack) to facilitate bidirectional cross-frame interaction.
The update interval of template is 20 and memory size is 50.
During inference, multiple templates are uniformly sampled from memory to interact with the search region.
We replace the feature extractor of SMTrack with Vision Transformer or BiSSM and concatenate multiple target templates into inputs for temporal modeling as previous trackers do~\cite{mixformerv2,yan2021learning,MambaVT}.
As shown in Table~\ref{tab:ablation_temporal}, $Trans_i$ and $BiSSM_i$ mean temporal modeling paradigms based on Transformer and BiSSM with $i$ target templates.
Both Transformer and BiSSM paradigms need numerous computational costs for temporal modeling during tracking.
Differently, our new temporal modeling paradigm of SMTrack builds interactions between the search region and temporal cues via hidden states without needing the involvement of template features, which releases computational costs of handling temporal cues.
Meanwhile, thanks to the multi-frame propagation strategy as described in Section~\ref{sec:imp_details}, SMTrack can flexibly utilize more temporal cues to discriminate the target object, thus achieving higher performance.

\noindent
\textbf{Analysis of the Hidden State Number}.
We take an ablation study with different hidden state numbers $L$ (state dimensions).
As shown in Table~\ref{tab:hidden_state}, SMTrack has a dramatic performance degradation with a small hidden state number $L=4$.
The performance of SMTrack increases as the number of states increases.
After the number of states reaches 16, the impact on performance flattens out.
The reason could be that 16 hidden states can capture enough temporal cues for target discrimination.
More hidden states may lead to repeated cues that are less computationally efficient and do not contribute to performance.
Combining performance and efficiency, we utilize 16 hidden states in SMTrack ($L=16$).

\begin{figure}[t]
    \centering
    \includegraphics[width=0.9\linewidth]{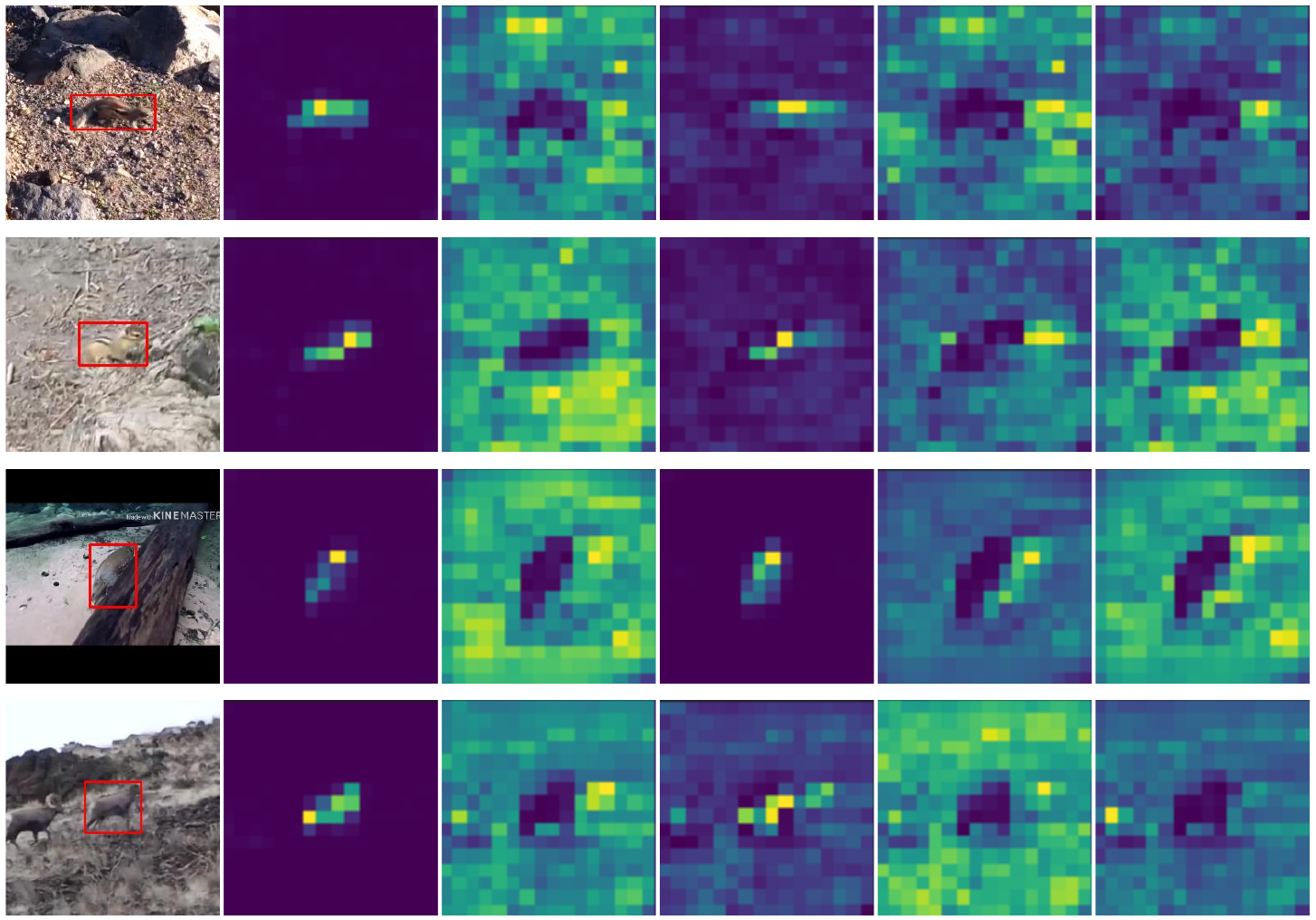}
    \caption{Visualization of SASM delta maps for different states. Different columns represent the delta map on the template of different hidden states. Our selective state-aware space model enables different states to adaptively capture diverse temporal cues. While previous SSMs ($e.g.$ Mamba) employ a state-shared timescale parameter, making features uniformly ignored or considered for all states updating, which limits the capability of hidden states to carry more diverse temporal cues.}
    \label{fig:sup_deltamap}
\end{figure}

\begin{figure}[t]
    \centering
    \includegraphics[width=0.9\linewidth]{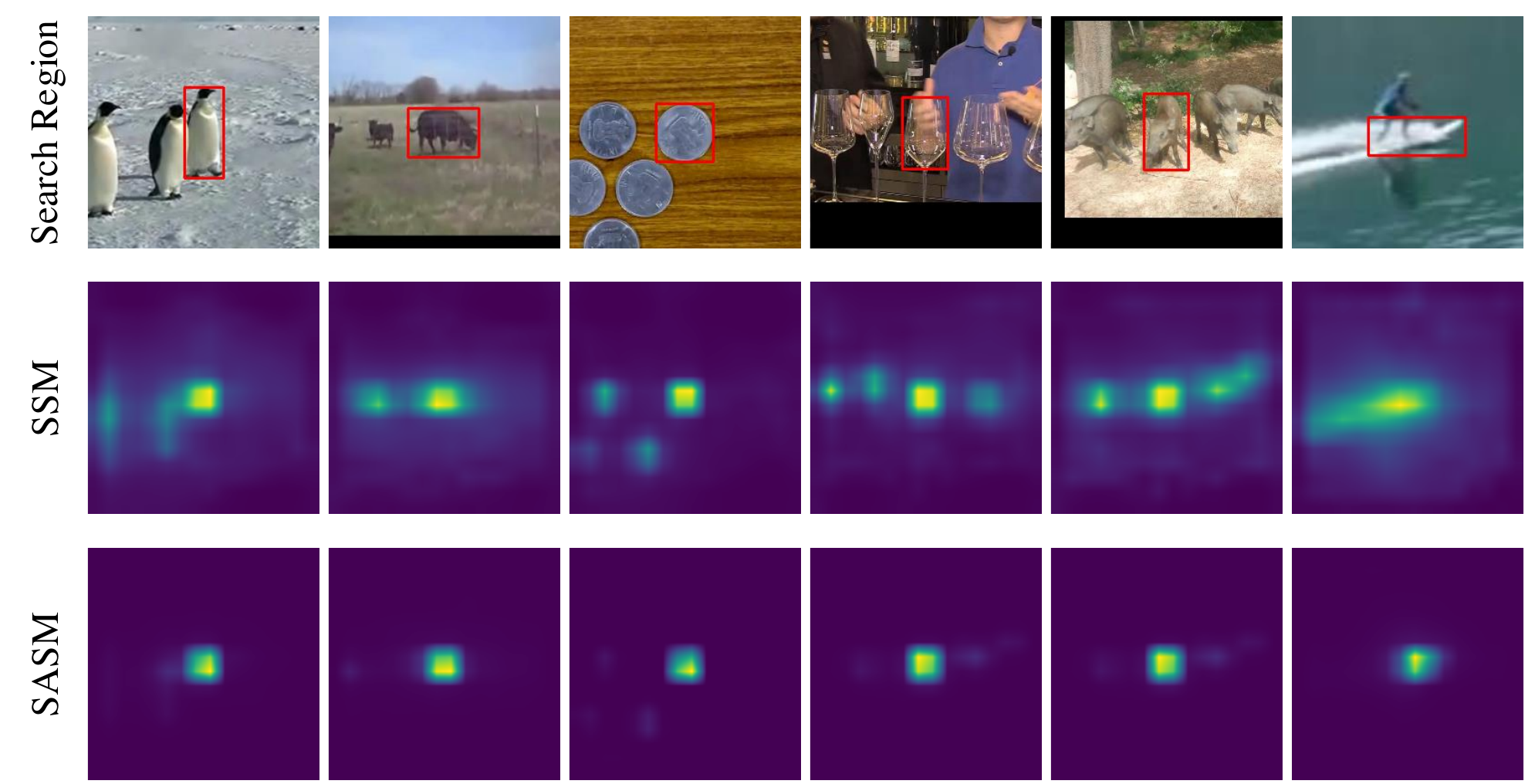}
    \caption{Visualization of classification score maps with SSM and SASM. Our selective state-aware space model can provide more robust classification score maps with less background response. This is because SASM can acquire more diverse temporal cues to not only localize the target object but also suppress backgrounds and distractors for robust tracking.}
    \label{fig:sup_clsmap}
\end{figure}

\noindent
\textbf{Analysis of the State Updating and Propagation}.
As shown in Table~\ref{tab:ablation_updating}, the updating interval has little effect on performance within a certain range.
Also, we try to integrate a template evaluator (TE) to select high-quality templates as Mixformer~\cite{cui2022mixformer} does.
Specifically, we first extract RoI features from the search region using the predicted box. 
A learnable score token is then introduced to perform attention interactions with both the RoI features and the template features. 
The final quality score is regressed through a multi-layer perceptron (MLP). 
The template evaluator is trained with balanced sampling of positive/negative templates. 
During tracking, templates scoring above 0.5 threshold are reserved for hidden states updating, ensuring robustness against template degradation.
We find that the template evaluator contributes little for SMTrack.
We can conclude that SMTrack is insensitive to the possible accumulation of errors, which benefits from the multi-frame propagation strategy described in Section~\ref{sec:imp_details}.
Therefore, SMTrack can avoid the design of the template evaluator, which will bring complex network architecture and training pipeline.
% 我们的时序线索的更新不需要额外模块的设计，如Iou、q
% Thanks to the linear form interactions between image features and hidden states in Equation~\ref{equ:4}, we can easily establish the interactions between the search region and multiple target templates via the sum of their hidden states without needing extra computational costs.
% %
% So, we sample multiple hidden states of previous templates to build multi-frame longer-range dependencies.
%
Also, we try different ways to sample multiple hidden states of previous templates for building multi-frame propagation.
As shown in Table~\ref{tab:ablation_propagation}, $Last_i$ means we sample hidden states of the last $i$ templates to interact with the search region, $Uni_i$ means we sample hidden states of $i$ scanned templates evenly with equal intervals according to their frame index as stated in Section~\ref{sec:imp_details}.
% 建立更大范围更多
We can find that introducing hidden states of more templates is generally beneficial for robust tracking.
Also, building interactions with longer-range temporal cues using uniform sampling can lead to significant improvements.

\noindent
\textbf{Analysis of the Memory Size}.
As shown in Table~\ref{tab:memory_size}, the performance gradually improves with increasing memory size, as larger memories preserve more diverse temporal hidden states that enhance discriminative capabilities.
We empirically set memory size $|\mathcal{M}|=50$ as this avoids excessive memory consumption while retaining sufficient temporal context for accurate predictions.

\subsection{Visualizations}
\label{sec:vis}

\noindent
\textbf{Delta Maps}.
Figure~\ref{fig:sup_deltamap} visualizes timescale parameters $\Delta$ of the target template for different states in SASM (we reshape $\Delta$ into 2D maps, termed delta maps).
The first column is the target template.
The previously widely used selective state space model~\cite{Mamba} employs a state-shared timescale parameter $\Delta$ as a gating mechanism, making features uniformly ignored or considered for all states updating.
Differently, as shown in Figure~\ref{fig:sup_deltamap}, our selective state-aware space model enables different states to adaptively capture diverse temporal cues ($e.g.$ targets, backgrounds, and distractors) in the target template via state-wise timescale parameters.
Thus, more diverse temporal cues are carried in hidden states, which facilitates SMTrack to not only localize the target object but also suppress backgrounds and distractors for robust tracking.

\begin{figure}
    \centering
    \includegraphics[width=0.9\linewidth]{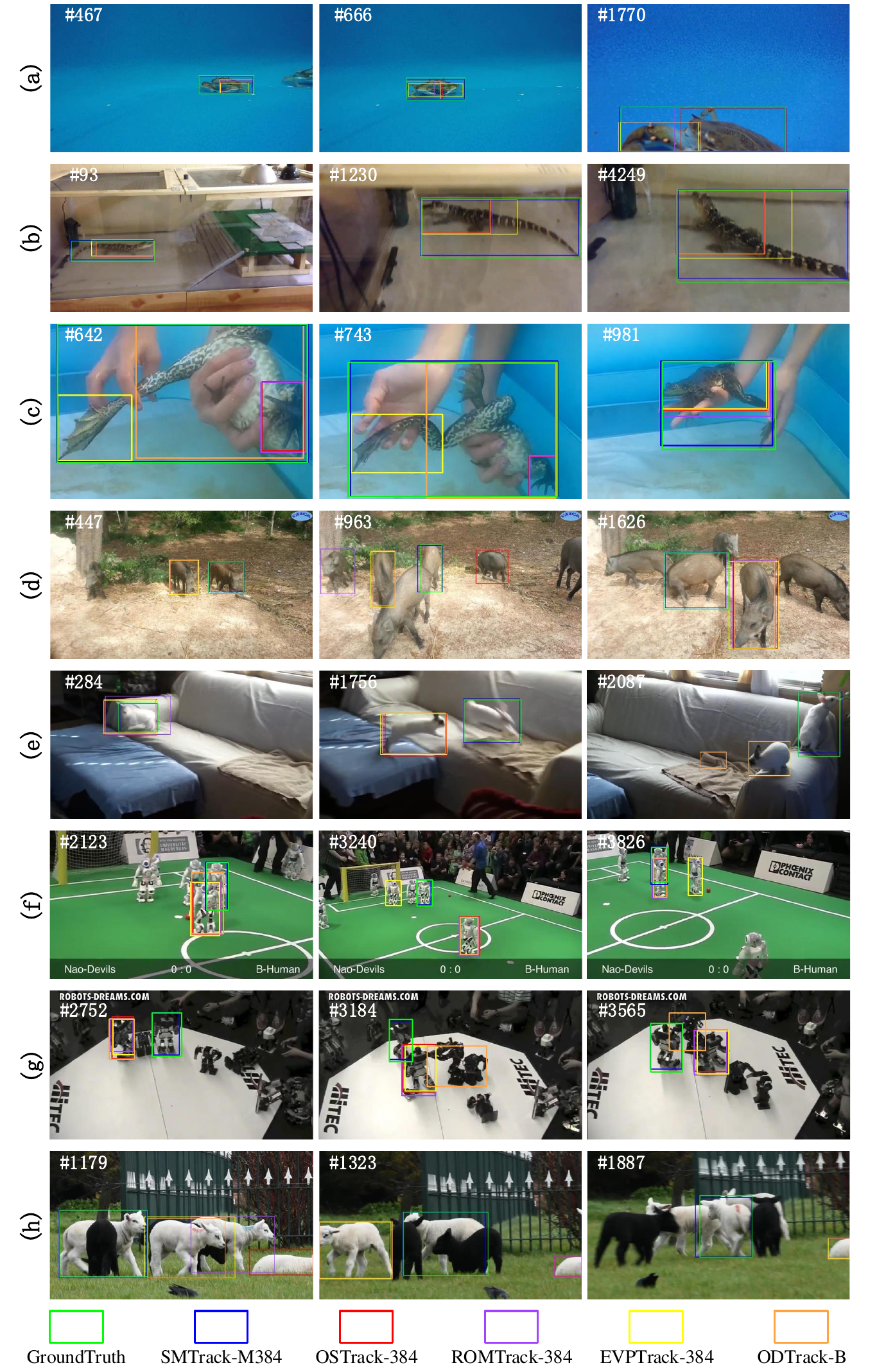}
    \caption{Visualization of tracking results in challenging scenarios. SMTrack achieves superior results compared with Transformer-based trackers with temporal modeling.}
    \label{fig:sup_visualization}
\end{figure}

\begin{figure}[t]
    \centering
    \includegraphics[width=0.85\linewidth]{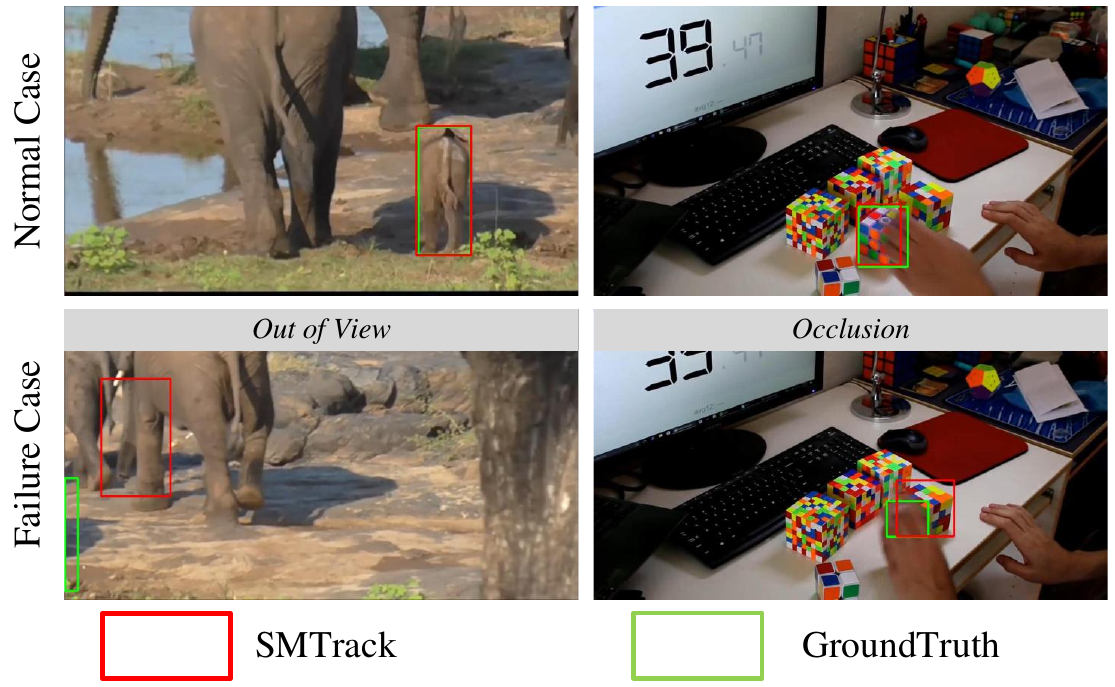}
    \caption{Visualization of SMTrack failure cases. The tracking performance of SMTrack is limited in scenarios involving out-of-view targets or occlusion.}
    \label{fig:failure}
\end{figure}

\noindent
\textbf{Classification Maps}.
As shown in Figure~\ref{fig:sup_clsmap}, we visualize the classification score map, which determine whether the tracker will drift to other objects and are vital for tracking robustness.
\textit{SSM} means that we utilize state-shared timescale parameters for hidden state learning as Mamba~\cite{Mamba} does.
\textit{SASM} means that we utilize our proposed state-wise timescale parameters for hidden state learning.
As we can see, our SASM can provide more robust classification score maps with less background response. 
This is because SASM can acquire more diverse temporal cues to not only localize the target object but also suppress backgrounds and distractors for robust tracking.

\noindent
\textbf{Tracking Result Comparison}.
As shown in Figure~\ref{fig:sup_visualization}, our SMTrack achieves more robust results in challenging dynamic scenarios, such as deformation (a,b,c), occlusion (c,f,h), distractor (d,f,g,h), motion blur (e).
Our SMTrack outperforms the previous popular Transformer-based tracker, OSTrack, with the same box head, showing the effectiveness of temporal modeling in complex dynamic scenarios.
Also, our SMTrack achieves superior results compared with Transformer-based trackers with temporal modeling.
Notably, our SMTrack eliminates the necessity for extra module designs or complicated training pipelines.
This is thanks to the designs of the SASM block, which enable temporal cues to be integrated via the updating and propagation of hidden states with little computational costs.
These results demonstrate the superiority of our novel temporal modeling paradigm.

\noindent
\textbf{Failure Cases}.
As shown in Figure~\ref{fig:failure}, the tracking performance of SMTrack is limited in scenarios involving out-of-view targets or occlusion.
Since SMTrack relies on appearance information for target localization, it is prone to tracking drift toward other objects in the background when the target features degrade in the tracking frame. 
We argue that further research into motion patterns of objects could provide additional discriminative cues to address these challenging scenarios.

%-------------------------------------------------------------------------
\section{Conclusion}

\begin{table}[t]
    \centering
    \caption{Performance and speed comparison on Trackingnet}
    \label{tab:speed}
    \resizebox{\linewidth}{!}{
    \begin{tabular}{c|ccccc}
    \hline
        \multirow{2}{*}{Method} & SeqTrack-B384 & ROMTrack-384 & EVPTrack-384 & ODTrack-B & SMTrack-M384 \\
        & \cite{seqtrack} & \cite{ROMTrack} & \cite{EVPTrack} & \cite{ODTrack} &  \\
    \hline
        AUC $\uparrow$ & 83.9 & 84.1 & 84.4 & 85.1 & \textcolor{red}{\textbf{85.2}} \\\hline
        GFlops $\downarrow$ & 148 & 91.5 & 78.9 & 86.2 & \textcolor{red}{\textbf{48.7}} \\
        FPS $\uparrow$ & 15 & 28 & 28 & 32 & \textcolor{red}{\textbf{34}} \\
    \hline
    \end{tabular}
    \vspace{-3mm}
    }
\end{table}

We propose a novel temporal modeling paradigm for visual tracking, SMTrack, providing a neat pipeline for training and tracking with long-range temporal cues.
SMTrack builds interaction between search region and temporal cues via hidden state propagation and updating without needing extra module designing or numerous computational costs.
We propose a selective state-aware space model (SASM) with state-wise timescale parameters to capture more diverse temporal cues.
As well as, we introduce interactions of hidden states to build dense dependencies between them for robust feature learning.
Extensive experimental results demonstrate the efficacy of SMTrack.

\subsection{Limitations}
The parallelization capability of our selective state-aware space model (SASM) is inferior to that of Transformer.
This is due to the fact that advanced state space models, including our proposed SASM and popular Mamba~\cite{Mamba}, do not utilize matrix multiplication units, which are specifically optimized by modern accelerators such as GPUs and TPUs.
This limits the advantage of SMTrack in speed comparisons with Transformer-based trackers on GPUs.
SMTrack-S256, SMTrack-M256 and SMTrack-M384 are run at 50 FPS, 36 FPS and 34 FPS respectively.
Nevertheless, the speed of SMTrack-M384 still exceeds that of popular trackers with auto-regressive modeling or consecutive temporal modeling, as shown in Table~\ref{tab:speed}.
We argue that the speed potential of SMTrack can be further unlocked through the optimization of hardware accelerators or hardware-friendly implementation in the future.

\subsection{Applications}
As an efficient temporal tracker, SMTrack achieves high tracking performance with low computational costs. 
It demonstrates strong potential for real-world applications in embedded devices such as drones and professional recording equipment (e.g., camera gimbals) to continuously estimate target object locations during automated tracking. 
SMTrack also exhibits broad applicability to deployment scenarios including robotics navigation and intelligent surveillance systems.

%-------------------------------------------------------------------------

\ifCLASSOPTIONcaptionsoff
  \newpage
\fi

% {
% \{mainbib,reference}
% }
% \clearpage
\bibliographystyle{IEEEtran}
\bibliography{mainbib}

\end{document}